\documentclass[10pt,twocolumn,letterpaper]{article}

\usepackage{cvpr}

\definecolor{cvprblue}{rgb}{0.21,0.49,0.74}
\usepackage[pagebackref,breaklinks,colorlinks,allcolors=cvprblue]{hyperref}

\usepackage{newfloat}
\usepackage{listings}
\usepackage{booktabs}
\usepackage{multirow}
\usepackage{graphicx}
\usepackage{makecell}
\usepackage{marvosym}

\usepackage[accsupp]{axessibility}

\def\paperID{42176} % *** Enter the Paper ID here
\def\confName{CVPR}
\def\confYear{2026}

\title{Scalable Object Relation Encoding for Better 3D Spatial Reasoning\\in Large Language Models}

\author{
Shengli Zhou$^{1}$\thanks{Work done as an intern at Peking University.~~\Letter~Corresponding author.} \quad Minghang Zheng$^2$ \quad Feng Zheng$^1$ \quad Yang Liu$^{2,3\text{\Letter}}$ \\
$^1$Department of Computer Science and Engineering, Southern University of Science and Technology\\
$^2$Wangxuan Institute of Computer Technology, Peking University\\
$^3$State Key Laboratory of General Artificial Intelligence, Peking University\\
{\tt\small zhousl2022@mail.sustech.edu.cn, \{minghang, yangliu\}@pku.edu.cn, f.zheng@ieee.org}
}

\begin{document}
\maketitle
\begin{abstract}
Spatial reasoning focuses on locating target objects based on spatial relations in 3D scenes, which plays a crucial role in developing intelligent embodied agents.
Due to the limited availability of 3D scene-language paired data, it is challenging to train models with strong reasoning ability from scratch.
Previous approaches have attempted to inject 3D scene representations into the input space of Large Language Models (LLMs) and leverage the pretrained comprehension and reasoning abilities for spatial reasoning.
However, models encoding absolute positions struggle to extract spatial relations from prematurely fused features, while methods explicitly encoding all spatial relations (which is quadratic in the number of objects) as input tokens suffer from poor scalability.
To address these limitations, we propose QuatRoPE, a novel positional embedding method with an input length that is linear to the number of objects, and explicitly calculates pairwise spatial relations through the dot product in attention layers.
QuatRoPE's holistic vector encoding of 3D coordinates guarantees a high degree of spatial consistency, maintaining fidelity to the scene's geometric integrity.
Additionally, we introduce the Isolated Gated RoPE Extension (IGRE), which effectively limits QuatRoPE's influence to object-related tokens, thereby minimizing interference with the LLM's existing positional embeddings and maintaining the LLM's original capabilities.
Extensive experiments demonstrate the effectiveness of our approaches.
The code and data are available at \href{https://github.com/oceanflowlab/QuatRoPE}{https://github.com/oceanflowlab/QuatRoPE}.
\end{abstract}

\begin{figure*}[t]
    \centering
    \includegraphics[width=0.98\textwidth]{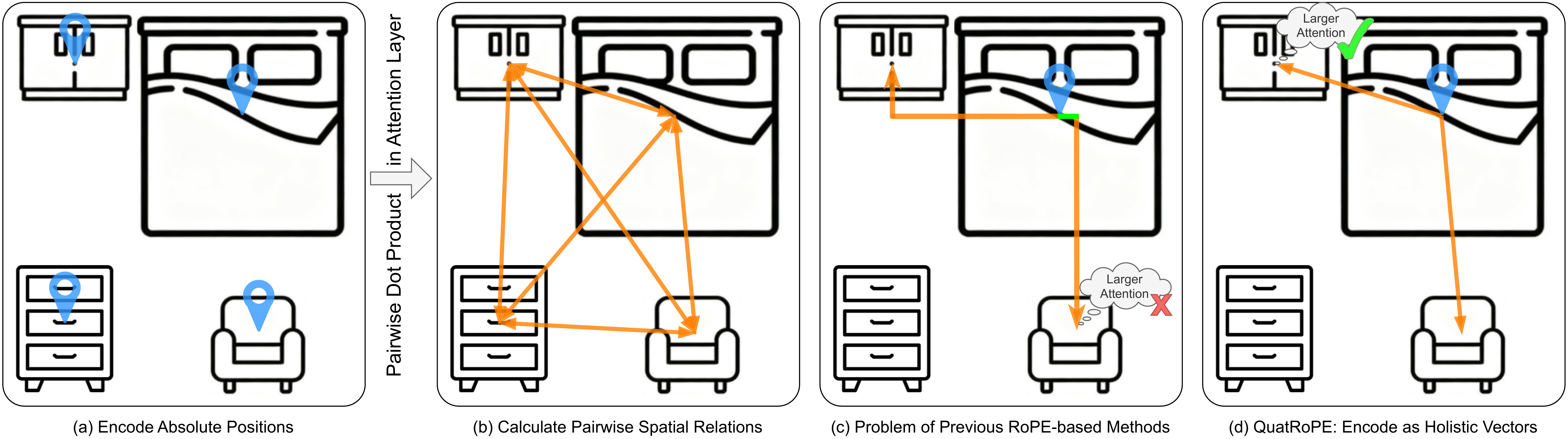}
    \caption{(a) In QuatRoPE, we embed the absolute 3D position of each object to the corresponding token, thus limiting the input length linear to object count. (b) By leveraging a dedicated rotation scheme, when tokens with 3D position embedding interact in the attention layer of the LLM, the absolute coordinates are transformed into pairwise relative positions, empowering spatial reasoning. (c) In previous methods, as the positions are decoupled into individual coordinates, when the coordinate of some axis is close (as marked in green), the attention score is incorrectly inflated. (d) QuatRoPE encodes positions as holistic vectors, correctly representing spatial relations.}
    \label{fig:teaser}
\end{figure*}    
\section{Introduction}

Spatial reasoning refers to the process of locating a target object according to its spatial relations with other objects (i.e., anchor objects) in the scene. Such a process is the core step for solving 3D Vision-Language (3D VL) tasks, including 3D Visual Grounding (3D VG) and 3D Visual Question-Answering (3D VQA). As the process of spatial reasoning is based on the spatial relations between objects, the accurate perception of inter-object spatial relations is a prerequisite for acquiring a strong spatial reasoning ability.
Thus, a core challenge in spatial reasoning is effectively encoding and computing object relations.

Due to the scarcity of 3D scene-text paired data, training a model with a strong spatial reasoning capability from scratch is challenging. With the development of Large Language Models (LLMs), previous works \cite{3dllm, chat-scene, 3dgraphllm} have integrated point cloud representations with natural language, leveraging LLMs' large-scale pretrained reasoning abilities to perform spatial reasoning on 3D scenes~\cite{mllm_survey_cje}.
In these works, the models represent scene layouts using either absolute or relative object positions.
\textbf{(1)} Absolute position encoding incorporates objects' 3D coordinates as part of their features~\cite{chat-scene,pq3d,scenellm}.
However, absolute coordinates carry little inherent meaning since the origin and orientation in 3D scenes have no natural physical definition, despite preserving geometric relationships between objects.
Moreover, since absolute positional encoding does not explicitly represent relative geometry, models must laboriously learn these relations from limited data. This challenge is further compounded by premature feature fusion, which obstructs LLMs from extracting positions and computing pairwise object relationships.
\textbf{(2)} For methods that directly encode pair-wise object relations using additional input tokens, the length of the LLM's input sequence grows quadratically with object count, which can easily exceed the input limits of many LLMs (e.g., the InteriorGS \cite{interiorgs} dataset contains an average of over 554 objects per scene, yielding over 153,181 relations).
While pruning strategies, such as 3DGraphLLM's \cite{3dgraphllm} KNN approach, reduce tokens by keeping only nearby objects, this risks omitting critical relations since spatial proximity does not ensure relevance, potentially causing errors in spatial reasoning.

In contrast to previous approaches, we propose \textbf{QuatRoPE}, which uses only $O(n)$ input tokens while preserving all $O(n^2)$ spatial relations (where $n$ is the number of objects in the scene), supporting scalability and avoiding erroneous pruning.
As shown in Fig. \ref{fig:teaser} (a) and (b), the core idea of QuatRoPE is to inject explicit absolute positional encodings for all object-related tokens\footnote{Object-related tokens: LLM's input tokens for objects' 2D/3D features and identifiers like \texttt{<obj001>}.} and leverage the Transformer's attention mechanism to \textbf{convert absolute positions into relative relationships} during query-key dot products.
Specifically, we apply quaternion rotations to query and key vectors based on the corresponding objects' 3D coordinates. By constructing specific mathematical formulations for rotation, the dot product (i.e., attention score) between two rotated vectors depends solely on their relative positions in the 3D scene, efficiently providing pairwise spatial relations for LLM.
Additionally, as shown in Fig. \ref{fig:teaser} (d), QuatRoPE encodes object coordinates as holistic vectors (instead of encoding the coordinate on each axis independently). Such an approach prevents inflated attention scores from small coordinate differences on single axes, accurately representing spatial layouts.

The scarcity of 3D scene-text paired data also makes it difficult to train LLMs with dual RoPE (i.e., language RoPE and QuatRoPE) from scratch. At the same time, when applying QuatRoPE on LLMs with Language RoPE, both RoPEs rotate query and key vectors, yielding interference and hindering position perception for both text and objects.

To address this issue, we further propose \textbf{Isolated Gated RoPE Extension (IGRE)}. In IGRE, object-related tokens are extended with QuatRoPE-specific dimensions (zero-padded for other tokens), isolating QuatRoPE from language RoPE. Also, IGRE ensures that attention scores only adjust to reflect relative positions when two object tokens interact through the dot product (i.e., gated), preserving the LLM's original linguistic capabilities.

While benchmarks like SQA3D~\cite{sqa3d}, ScanRefer~\cite{scanrefer}, and Multi3DRef~\cite{multi3dref} evaluate aspects of spatial understanding, they are not designed to—and thus are inherently limited in—purely assessing spatial reasoning.
In these tasks, language descriptions often intertwine spatial relationships with non-spatial cues, such as object categories or attributes. This makes it difficult to determine whether a model's success stems from true spatial comprehension or from simply recognizing semantic or visual features. To address this deficiency, we introduce a diagnostic benchmark, the Attribute-free Spatial Reasoning (\textbf{ASR}) benchmark, to isolate and more directly probe a model's spatial reasoning capabilities.
In our benchmark, we select ScanQA's \cite{scanqa} uniquely-answerable object-name questions, filter out those revealing target attributes to enforce spatial reasoning, and convert them into 3D VG format to eliminate language generation biases. By such an approach, the ASR benchmark can make a fair and rigorous comparison of spatial reasoning.
Across all these benchmarks, our approach consistently outperforms strong baselines, showing that QuatRoPE provides effective positional cues for spatial understanding.

In summary, our contributions are as follows: (1) We propose \textbf{QuatRoPE}, a novel 3D positional encoding that explicitly models objects' pairwise relative positions through quaternion rotations, enhancing the spatial understanding for 3D LLMs. (2) We propose \textbf{IGRE} to combine QuatRoPE with the language RoPE to reduce interference. (3) We construct a challenging benchmark \textbf{ASR} for exclusively evaluating 3D spatial understanding. (4) We achieve consistent and large-margin gains on ASR and multiple existing 3D VL benchmarks, validating the effectiveness.
\section{Related Work}

\subsection{3D VL Tasks on Spatial Reasoning}

3D Vision-Language (3D VL) refers to multi-modal tasks that are solved by combining 3D scenes and natural language, such as 3D Visual Grounding (3D VG) \cite{Ground_2025_ICRA} and 3D Visual Question-Answering (3D VQA) \cite{aqua,vqa_cje,qa_cje}.

Previously, ScanRefer \cite{scanrefer} introduced the task of single-object 3D VG, where the model finds an object based on a text description (e.g., locating ``the bottle on top of the table''); Multi3DRef \cite{multi3dref} extended this task to cases where the number of ground-truth objects varies, further testing the model's spatial reasoning skills. For 3D VQA, ScanQA \cite{scanqa} was the first to define the task of answering questions about 3D scenes (e.g., answering ``What color is the object under the chair and next to the lamp?''). SQA3D \cite{sqa3d} further developed this into situated question-answering, where models answer questions from a specific viewpoint, which better aligns with the practical requirements for applications such as intelligent robots. Other datasets, such as Nr3D and Sr3D \cite{referit3d}, have also defined variants of these 3D VL tasks.

However, current benchmarks in these tasks fail to directly reflect models' spatial reasoning ability, as objects' attributes (e.g., category, color, and shape) in language descriptions can help models locate target objects without spatial reasoning. In contrast, we propose a diagnostic benchmark that omits all attributes of the target objects, thereby evaluating models' spatial reasoning abilities exclusively.

\subsection{3D LLMs for Spatial Reasoning}

When solving spatial reasoning tasks, models should be able to precisely perceive the spatial relation between objects to obtain the correct answer.
Due to the scarcity of 3D scene-text paired data, previous works have leveraged the perception and reasoning capabilities of LLMs to enhance spatial reasoning.
Among these works, 3D-LLM \cite{3dllm} represents the 3D entire scene as a holistic feature. Though such an encoding approach can preserve the scene layout, the compact representation loses details and entangles objects' features, which requires the model to identify objects and impedes object-level spatial reasoning.

To facilitate object-level reasoning, LEO \cite{leo} and Chat-Scene \cite{chat-scene} segment the scene into objects and encode the feature of each object as input tokens. 
Despite their promising performance, they struggle to extract spatial relations between objects from absolute positions that are prematurely fused with geometric features.
To solve this problem, 3DGraphLLM \cite{3dgraphllm} utilizes additional input tokens to explicitly represent spatial relations between objects. Additionally, since the number of relations is quadratic to the object count (which can easily exceed LLMs' input limits), 3DGraphLLM employs a K-Nearest-Neighbors (KNN) strategy, encoding only the spatial relations between each object and its nearest objects. However, this approach is error-prone as proximity does not always indicate task-relevant importance.

In contrast, we propose QuatRoPE that encodes 3D positions on each object-related token. Through a dedicated embedding scheme, it converts absolute coordinates to pairwise relative positions of all objects via query-key dot products in attention layers. Such a method not only ensures robustness to global rotations and translations but also mitigates error-prone pruning.

\subsection{Rotary Positional Embeddings}

Rotary Positional Embedding (RoPE) \cite{rope} enhances transformers by encoding relative positions through complex rotations of query/key vector segments of 2 components. Each segment is rotated by $m\theta_i$ (where $m$ is the absolute position and $\theta_i$ is the frequency), making attention scores depend only on position differences. 
By this mechanism, the dot products of query and key vectors are only related to the difference in position, transforming absolute positions into relative positions.
Currently, RoPE has become foundational in various LLMs, including LLaMA \cite{llama3} and QWen \cite{qwen}.

For multi-modal data (e.g., images), M-RoPE \cite{qwen2-vl} extends RoPE by grouping segments for multi-position embedding.
Video-RoPE \cite{videorope} further introduces Low-frequency Temporal Allocation to focus on long-range dependencies along the time axis, and Diagonal Layout to maintain spatial-textual position consistency.

However, 3D scenes pose unique challenges: existing methods overemphasize proximity along individual axes. When two objects have similar coordinates on one axis (despite being distant overall), these methods inflate attention scores due to incorrectly amplified dimension-wise products in segment groups corresponding to the axis. This creates false ``nearby'' associations between objects, impairing the model's understanding of spatial relationships.
In contrast, QuatRoPE encodes coordinates as integrated vectors by rotating each individual dimension of the query and key vectors according to the corresponding 3D coordinates. This approach ensures attention scores increase only when objects are truly proximate in 3D space, effectively representing scene layouts.
\section{Method}

\subsection{Baseline Models Revisited}

To utilize LLMs for perceiving and reasoning on scene information, previous 3D LLMs \cite{chat-scene,3dgraphllm} have aligned and injected point cloud features into LLMs. In these works, the pipeline is as follows:

The model is input with a point cloud and textual instructions. To begin with, the model utilizes either ground-truth segmentations or predictions from off-the-shelf segmentation models \cite{mask3d,seg_cje,DA_2025_IROS} to segment the point cloud into a series of objects, thereby facilitating the model's ability to perform object-level reasoning.
For each object, its features (e.g., 3D geometric feature calculated by PointNet++ \cite{pointnetpp}) are projected into the input space of LLM through projection layers.
Additionally, a set of object identifiers is defined and trained to fit within the input space of LLMs (e.g., \verb|<obj005>| represents the fifth object in the scene enumeration order), helping the model refer to specific objects in the scene.
Finally, the project features and object identifiers are parsed into LLMs as input tokens (each object-related token corresponds to a single object), along with other language tokens for prompts and questions.

Inside the LLM, the feature vectors of tokens serve as the input embeddings for the first self-attention layer. Thus, when the model calculates attention scores between tokens, it also forms attention associations between objects based on their features.
In previous works that prematurely fuse absolute coordinates into objects' overall features, the spatial information in feature vectors is implicit and sparse, weakening the relative spatial information in the association.
Therefore, in QuatRoPE, we enhance the spatial information by providing an explicit encoding on each object-related token, representing its absolute position in the scene. When object-related tokens interact in the attention layers of the LLM, the absolute positions can be further transformed into relative spatial cues between objects, empowering the model's understanding of spatial relations.

\subsection{QuatRoPE}

To provide the LLM with pairwise spatial relations between objects, while constraining the number of input tokens to be linear to the number of objects, we propose QuatRoPE. The core mechanism of QuatRoPE is to first encode the corresponding object's absolute coordinates on object-related tokens, and then calculate pairwise relative positions between objects during the dot products for query and key vectors in attention layers.

Initially, given that spatial reasoning operates at the object level, we represent each object's 3D position through its bounding box center.

To facilitate relative-position calculation via dot products in attention layers, we encode absolute positions using rotations. Such an approach transforms absolute coordinates into relative positions, since dot products reflect angle differences.
Specifically, the query vectors and the key vectors in self-attention layers are grouped into 3D segments, represented each as a pure quaternion (denoted as $\vec q$ and $\vec k$ with zero real part), and apply quaternion rotation before each attention layer.
Let $\vec m$ and $\vec n$ be the absolute 3D coordinates of the objects corresponding to query vector $\vec q$ and key vector $\vec k$, and $f(\vec x, \vec p)$ be the function for rotating the query or key vector $\vec x$ according to the corresponding absolute 3D position $\vec p$.
Since the attention score should only relate to the relative position (i.e., $\vec m-\vec n$), the rotation function $f$ should satisfy for some ternary function $g$:

\begin{equation} \label{eq:quat_goal}
    \left<f(\vec q, \vec m), f(\vec k, \vec n)\right> = g(\vec q, \vec k, \vec m-\vec n)
\end{equation}

When encoding coordinates independently, the proximity of coordinates on a single axis can mislead the model into amplifying attention scores for objects that are indeed far away. Thus, QuatRoPE embeds the coordinates as a unified vector, i.e., the components of the query and key vectors are adjusted based on the object's position rather than its coordinate along a specific axis. Since coordinates are 3D vectors, we leverage quaternion rotation with three degrees of freedom to embed them. Formally, the rotation function can be expressed via Euler angle decomposition as:

\begin{equation} \label{eq:quat_repr}
    \begin{cases}
        f(\vec q, \vec m)=Q(\vec m)~\vec q~Q^{-1}(\vec m)\\
        Q(\vec m)=Q_z(m_z)~Q_y(m_y)~Q_x(m_x)\\
        Q_x(m_x)=\cos\left[\theta_x(m_x)/2\right]+\hat i\sin\left[\theta_x(m_x)/2\right]\\
        Q_y(m_y)=\cos\left[\theta_y(m_y)/2\right]+\hat j\sin\left[\theta_y(m_y)/2\right]\\
        Q_z(m_z)=\cos\left[\theta_z(m_z)/2\right]+\hat k\sin\left[\theta_z(m_z)/2\right]
    \end{cases}
\end{equation}

\noindent where $Q$'s are rotation matrices and $\theta$'s are unary functions.

Through Equation (\ref{eq:quat_repr}), we transform the requirement in QuatRoPE (i.e., converting absolute coordinates to relative positions via dot products) into deriving $\theta$'s that satisfy Equation (\ref{eq:quat_goal}).
To solve the equation, we transform the dot product into the real part of the product of the rotation functions to yield a form with multiplication between the rotation matrices (i.e., $Q^{-1}(\vec m)$ and $Q(\vec n)$). 

\begin{equation} \label{eq:quat_exp}
    \begin{aligned}
        \left<f(\vec q, \vec m), f(\vec k, \vec n)\right>=&\Re[f(\vec q, \vec m)f^*(\vec k, \vec n)]\\
        =&\Re[Q(\vec m)~\vec q~Q^{-1}(\vec m)~Q(\vec n)~\vec k^*~Q^{-1}(\vec n)]
    \end{aligned}
\end{equation}

\noindent where $\vec k^*$ denotes the conjugate of quaternion $\vec k$, and $\Re$ denotes the real part of the quaternion. To pair every $Q(\vec m)$ with $Q(\vec n)$, according to the real-part invariance of quaternion rotation (i.e., $\Re(Q^{-1}kQ)=\Re(k)$), left multiplying $Q^{-1}(\vec m)$ and right multiplying $Q(\vec m)$ yields:

\begin{equation} \label{eq:quat_rot_inv}
    \left<f(\vec q, \vec m), f(\vec k, \vec n)\right>=\Re[\vec q~Q^{-1}(\vec m)~Q(\vec n)~\vec k^*~Q^{-1}(\vec n)~Q(\vec m)]
\end{equation}

According to Equation (\ref{eq:quat_goal}), since $\left<f(\vec q, \vec m), f(\vec k, \vec n)\right>$ should only relate to $\vec m-\vec n$, we have $Q^{-1}(\vec n)~Q(\vec m)=Q(\vec m-\vec n)$. The equation further yields that unary functions $\theta_x, \theta_y$, and $\theta_z$ should be linear (as detailed in the appendix).
Thus, an approximate solution for QuatRoPE is:

\begin{equation} \label{eq:quat_final}
    \begin{cases}
        f(\vec q, \vec m)=Q(\vec m)~\vec q~Q^{-1}(\vec m)\\
        Q(\vec m)=Q_z(m_z)~Q_y(m_y)~Q_x(m_x)\\
        Q_x(m_x)=\cos[m_x\theta_x(1)/2]+\hat i\sin[m_x\theta_x(1)/2]\\
        Q_y(m_y)=\cos[m_y\theta_y(1)/2]+\hat j\sin[m_y\theta_y(1)/2]\\
        Q_z(m_z)=\cos[m_z\theta_z(1)/2]+\hat k\sin[m_z\theta_z(1)/2]
    \end{cases}
\end{equation}

\noindent where $\theta_x(1)$, $\theta_y(1)$, and $\theta_z(1)$ are frequencies for quaternion rotations.
According to Equation (\ref{eq:quat_goal}), as we perform rotation by $\vec q:=f(\vec q,\vec m)$ and $\vec k:=f(\vec k, \vec n)$ before each attention layer, the attention scores between object-related tokens reflect their relative positions. By such an approach, QuatRoPE can effectively convey relative positional information for LLMs to perform spatial reasoning.

Moreover, for objects that are spatially close in a scene, their QuatRoPE embeddings are similar, resulting in larger attention scores. This behavior aligns with the human cognitive mechanism of Maxim of Relation \cite{logic_and_conv}. For example, when referring to ``the window to the left of the door,'' if multiple windows exist at varying distances, humans typically imply the one closest to the door. Such alignment consequently enhances the LLM's ability to comprehend implicit references in natural language.

\subsection{Isolated Gated RoPE Extension}

Although QuatRoPE can effectively provide spatial information for LLMs to utilize, training point cloud-based 3D LLMs presents new challenges. Due to the scarcity of 3D scene-text paired data, training an LLM with language RoPE and QuatRoPE from scratch is impractical. However, simply applying QuatRoPE along with language RoPE may cause interference as they simultaneously perform rotation on query and key vectors.

\begin{table*}[t]
    \centering
    \caption{Results for the comparative experiments, the best scores obtained by using ground-truth or predicted segmentation are underlined, and the overall best scores are in bold. By applying QuatRoPE, our models have achieved consistent gains under all metrics.} \label{tab:comp}
    \scalebox{0.79}{
    \begin{tabular}{l|c|cccc|cc|c}
    \toprule
        \multirow{2}{*}{\textbf{Model}} & \textbf{Detector /} & \multicolumn{4}{c|}{\textbf{ScanRefer}} & \multicolumn{2}{c|}{\textbf{Multi3DRef}} & \textbf{SQA3D} \\ 
        ~ & \textbf{Segmentation}& Acc@0.25 & Acc@0.5 & Multi@0.25 & Multi@0.5 & F1@0.25 & F1@0.5 & EM@1 \\ \midrule
        ScanRefer~\cite{scanrefer} & VoteNet & 39.0  & 26.1 & 32.1 & 21.3   & -- & -- & -- \\ 
        3DJCG~\cite{3djcg} & VoteNet & 49.6  & 37.3 & 41.4 & 30.8  & -- & 26.6  & -- \\ 
        Vil3DRef~\cite{vil3dref} & PointGroup & 47.9 & 37.7 & 40.3 & 30.7 & -- & -- & -- \\
        D3Net~\cite{d3net} & PointGroup & -- & 37.9  & -- & 30.1  & -- & 32.2  & -- \\ 
        VPP-Net~\cite{vppnet} & Group-free & 55.7  & 43.3 & 50.3 & 39.0  & -- & -- & -- \\ 
        AugRefer~\cite{augrefer} & Group-free & 55.7  & 44.0 & 50.0 & 39.1  & -- & -- & -- \\ 
        M3DRef-CLIP~\cite{multi3dref} & PointGroup & -- & 44.7 & -- & 36.8  & 42.8  & 38.4  & -- \\ 
        MA2TransVG~\cite{ma2transvg} & Group-free & 57.9  & 45.7 & 53.8 & 41.4  & -- & -- & -- \\ 
        3D-VisTA~\cite{3dvista} & Mask3D & 50.6  & 45.8 & 43.7 & 39.1  & -- & -- & 48.5  \\ 
        3DSyn~\cite{3dsyn} & Mask3D & 52.3  & 46.2  & -- & -- & -- & -- & -- \\ 
        TSP3D~\cite{tsp3d} & N/A & 56.5  & 46.7  & -- & -- & -- & -- & -- \\ 
        PQ3D~\cite{pq3d} & PQ3D Promptable & --  & 51.2  & -- & 46.2 & -- & 50.1 & 47.1 \\ 
        BridgeQA~\cite{bridgeqa} & VoteNet & -- & -- & -- & -- & -- & -- & 52.9  \\ 
        Scene-LLM~\cite{scenellm} & N/A & -- & -- & -- & -- & -- & -- & 53.6  \\ \midrule
        Chat-Scene-1B~\cite{chat-scene} & GT & 50.7  & 50.3  & 42.7  & 42.3  & 53.3 & 52.9 & 50.7  \\ 
        \textbf{Chat-Scene-1B + QuatRoPE (Ours)} & GT & 55.4  & 55.0  & 47.8  & 47.4  & 58.1 & 57.7 & 53.1 \\ 
        3DGraphLLM-1B~\cite{3dgraphllm} & GT & 55.9  & 55.8  & 47.9  & 47.7  & 58.6 & 58.4 & 51.1  \\ 
        \textbf{3DGraphLLM-1B + QuatRoPE (Ours)} & GT & \underline{\textbf{58.3}}  & \underline{\textbf{58.2}}  & \underline{50.8}  & \underline{\textbf{50.6}}  & \underline{\textbf{60.7}} & \underline{\textbf{60.5}} & \underline{53.2} \\ \midrule
        Chat-Scene-7B~\cite{chat-scene} & Mask3D & 55.5  & 50.2  & 47.8 & 42.9 & 57.1  & 52.4  & 54.6  \\ 
        \textbf{Chat-Scene-7B + QuatRoPE (Ours)} & Mask3D & 57.8  & 52.2  & 51.1 & 45.7  & 59.5  & 54.8  & 54.7  \\ 
        3DGraphLLM-7B~\cite{3dgraphllm} & Mask3D & 57.0  & 51.3  & -- & -- & 60.1  & 55.4  & 53.1  \\ 
        \textbf{3DGraphLLM-7B + QuatRoPE (Ours)} & Mask3D & \underline{58.2}  & \underline{52.5}  & \underline{\textbf{54.3}} & \underline{49.2} & \underline{60.6}  & \underline{56.0}  & \underline{\textbf{55.2}} \\ \bottomrule
    \end{tabular}
    }
\end{table*}

\begin{table*}[t]
    \centering
    \caption{Results on our spatial reasoning benchmark. Our model achieves significant and consistent gains across various settings.} \label{tab:sr_bench}
    \scalebox{0.85}{
    \begin{tabular}{cc|cccc}
    \midrule
        Model & LLM & Acc @ 0.25 & Gain & Acc @ 0.5 & Gain \\ \midrule
        Chat-Scene~\cite{chat-scene} & Llama-3.2-1B-Instruct & 22.92  & -- & 22.92  & -- \\ 
        \textbf{Chat-Scene + QuatRoPE (Ours)} & Llama-3.2-1B-Instruct & 27.38  & 4.46 (19.48\%)  & 27.38  & 4.46 (19.48\%)  \\ \midrule
        3DGraphLLM~\cite{3dgraphllm} & Llama-3.2-1B-Instruct & 25.89  & -- & 25.60  & -- \\ 
        \textbf{3DGraphLLM + QuatRoPE (Ours)} & Llama-3.2-1B-Instruct & 29.76  & 3.87 (14.94\%)  & 29.76  & 4.17 (16.28\%) \\ \midrule
        3DGraphLLM~\cite{3dgraphllm} & Llama-3-8B-Instruct & 37.50  & -- & 36.90  & -- \\ 
        \textbf{3DGraphLLM + QuatRoPE (Ours)} & Llama-3-8B-Instruct & 41.96  & 4.46 (11.90\%)  & 41.96  & 5.06 (13.71\%) \\ \midrule
    \end{tabular}
    }
\end{table*}

\begin{table*}[t]
    \centering
    \small
    \caption{Results for ablation study on different composition approaches, the best scores of each baseline are marked in bold. } \label{tab:abl_comp}
    \begin{tabular}{c|cccc|c}
    \toprule
        \multirow{2}{*}{\textbf{RoPE Composition Approach}} & \multicolumn{4}{c|}{\textbf{ScanRefer}} & \textbf{SQA3D} \\ 
        ~ & Acc @ 0.25 & Acc @ 0.5 & Multi @ 0.25 & Multi @ 0.5 & EM @ 1 \\ \midrule
        \multicolumn{6}{l}{Baseline: Chat-Scene~\cite{chat-scene}} \\ \midrule
        None & 50.72  & 50.33  & 42.69  & 42.29  & 50.72 \\ 
        Trans-Additive & 53.12  & 52.79  & 45.48  & 45.14  & 52.96 \\ 
        \textbf{IGRE (Ours)} & \textbf{55.44}  & \textbf{55.00}  & \textbf{47.81}  & \textbf{47.36}  & \textbf{53.14} \\ \midrule
        \multicolumn{6}{l}{Baseline: 3DGraphLLM~\cite{3dgraphllm}} \\ \midrule
        None & 55.92  & 55.75  & 47.92  & 47.74  & 51.09 \\ 
        Trans-Additive & 53.68  & 53.38  & 45.94  & 45.64  & 51.55 \\ 
        \textbf{IGRE (Ours)} & \textbf{58.30}  & \textbf{58.15}  & \textbf{50.77}  & \textbf{50.60}  & \textbf{53.20} \\ \bottomrule
    \end{tabular}
\end{table*}

\begin{table*}[t]
    \centering
    \small
    \caption{Results for ablation study on different RoPE methods, the best scores of each baseline are marked in bold.} \label{tab:abl_rope}
    \begin{tabular}{c|cccc|c}
    \toprule
        \multirow{2}{*}{\textbf{Explicit Positional Encoding Approach}} & \multicolumn{4}{c|}{\textbf{ScanRefer}} & \textbf{SQA3D} \\ 
        ~ & Acc @ 0.25 & Acc @ 0.5 & Multi @ 0.25 & Multi @ 0.5 & EM @ 1 \\ \midrule
        \multicolumn{6}{l}{Baseline: Chat-Scene~\cite{chat-scene}} \\ \midrule
        None & 50.72  & 50.33  & 42.69  & 42.29  & 50.72 \\
        Raw Coordinates & 52.26 & 52.01 & 44.41 & 44.17 & 51.40 \\ 
        M-RoPE & 54.30  & 53.92  & 46.44  & 46.10  & 51.55 \\
        \textbf{QuatRoPE (Ours)} & \textbf{55.44}  & \textbf{55.00}  & \textbf{47.81}  & \textbf{47.36}  & \textbf{53.14} \\ \midrule
        \multicolumn{6}{l}{Baseline: 3DGraphLLM~\cite{3dgraphllm}} \\ \midrule
        None & 55.92  & 55.75  & 47.92  & 47.74  & 51.09 \\
        Raw Coordinates & 3.60 & 3.44 & 3.57 & 3.46 & 35.50 \\ 
        M-RoPE & 57.69  & 57.48  & 50.07  & 49.83  & 53.14 \\ 
        \textbf{QuatRoPE (Ours)} & \textbf{58.30}  & \textbf{58.15}  & \textbf{50.77}  & \textbf{50.60}  & \textbf{53.20} \\ \bottomrule
    \end{tabular}
\end{table*}

Meanwhile, directly applying QuatRoPE rotations to query and key vectors also introduces erroneous associations between object-related tokens and non-object tokens (e.g., tokens for system prompts, questions, instructions, and relations). While RoPE-based positional encodings can represent arbitrary positions or coordinates, they inherently cannot express the concept that ``non-object tokens do not correspond to a position in the 3D coordinate system.'' Even if non-object tokens are left unrotated, it is equivalent to positioning them at $(0, 0, 0)$. Such a configuration misleadingly biases the model to disproportionately attend to relationships between non-object tokens and objects near the coordinate origin.

\begin{figure}[t]
    \centering
    \includegraphics[width=0.43\textwidth]{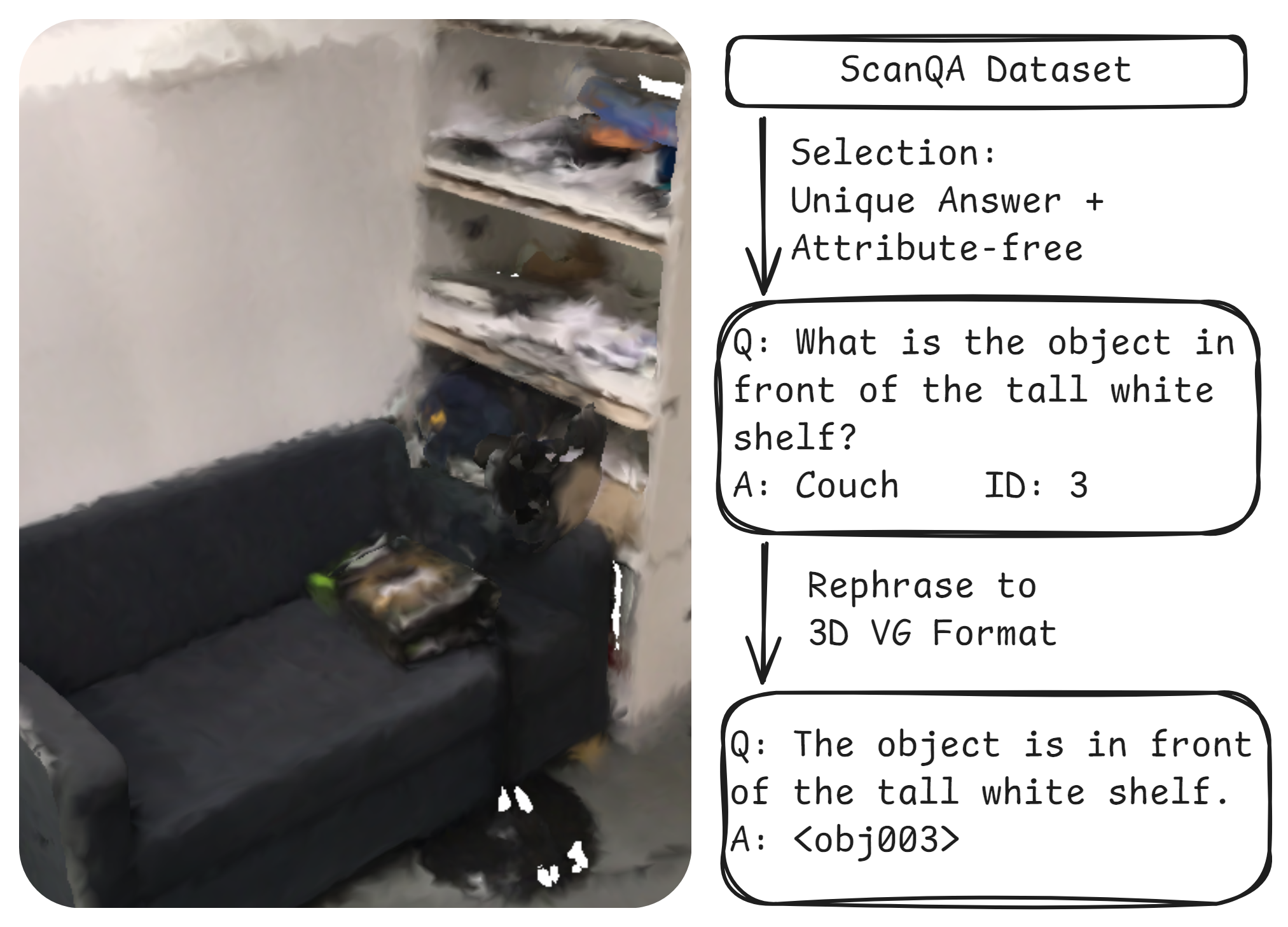}
    \caption{An illustration of ASR's construction pipeline.}
    \label{fig:asr}
\end{figure}

To address these problems, we introduce \textbf{Isolated Gated RoPE Extension (IGRE)}.
For an object-related token, we apply QuatRoPE on a base vector and concatenate it to the query/key vector.
For a non-object token, we concatenate a zero vector to pad the query/key vector to the same dimension as object-related tokens.

By this approach, we isolate the components rotated by language RoPE and QuatRoPE, thus reducing the interference between multiple RoPEs.
Additionally, as non-object tokens are zero-padded, the ``non-existence'' of non-object tokens in the 3D scene can be well-represented.
Also, under such a representation, when a query or key vector that belongs to a non-object token is involved in the dot product, the padded zeros ensure that element-wise products in these dimensions are 0, keeping the attention scores unchanged. Thus, IGRE can constrain QuatRoPE's adjustments on attention scores between object-related tokens, maximizing the retention of the pretrained LLM's capabilities in understanding natural language and performing reasoning.
QuatRoPE's adjustments to attention scores are gated within the dot products with both query and key vectors from object-related tokens. Therefore, IGRE can maximally reduce interference and preserve LLM's ability to understand natural language and perform reasoning.

\subsection{Attribute-free Spatial Reasoning Benchmark}

Though previous datasets in 3D VL tasks can reflect models' spatial reasoning abilities, none of them can fully isolate the impact of other model abilities on the final scores.
Under 3D VL tasks, descriptions of object attributes (e.g., category, color, and shape) often entangle with those for spatial relations, improperly facilitating the model an unintended bypass of spatial reasoning.
For example, for a 3D VG task locating ``the red chair under the window and next to the table'' while there is only one red chair in the scene, the model can obtain the answer by recognizing the red chair, rather than locating it through its relations with the window and the table.

To address these problems, we propose the Attribute-free Spatial Reasoning (\textbf{ASR}) benchmark.
First, we select a series of 3D VQA questions with unique answers in ScanQA that ask for the name of the object.
Then, we filter out questions that do not provide any other attributes of the target object, requiring the model to obtain the answer through spatial reasoning (e.g., ``What is the object in front of the tall white shelf?'').
Finally, we convert these queries into a 3D VG format (e.g., ``The object in front of the tall white shelf''), where the model only needs to perform multiple-choice selection between objects in the scene, eliminating the impact of different language generation abilities between models.
The pipeline for constructing our ASR benchmark is illustrated in Fig. \ref{fig:asr}.

Through attribute-free questions and the 3D VG format setting of our benchmark, we ensure fair and rigorous comparisons of models' spatial reasoning capabilities.
\section{Experiments}

\subsection{Experimental Settings}

We validate the effectiveness of our method through experiments using strong point cloud-based 3D LLM Chat-Scene \cite{chat-scene} and model 3DGraphLLM \cite{3dgraphllm} as baselines. All models are trained on a combined dataset composed of ScanRefer \cite{scanrefer}, Multi3DRef \cite{multi3dref}, ScanQA \cite{scanqa}, SQA3D \cite{sqa3d}, Scan2Cap \cite{scan2cap}, ReferIt3D \cite{referit3d}, and Chat-Scene's object alignment task. During training, LLMs are fine-tuned with LoRA (rank $r=16$ and scaling factor $\alpha=16$) at a learning rate of $2\times 10^{-5}$. For the 3DGraphLLM baseline, we adopt the same setting, pruning scene graphs using KNN with $k=2$. To evaluate spatial reasoning ability, we test models on 3D VG benchmarks, including ScanRefer and Multi3DRef (as they involve precise perception of objects' spatial relations and locating objects according to instructions), and Situated 3D VQA benchmark SQA3D.

\subsection{Comparative Experiments}

In this experiment, we aim to verify the effectiveness and generalizability of the proposed methods. We utilize Llama-3.2-1B-Instruct and Vicuna-7B-v1.5 as the LLM for Chat-Scene and 3DGraphLLM baselines, and apply QuatRoPE through IGRE to these models. Finally, we compare their performance with specialist and generalist models on multiple datasets to evaluate the gain for spatial reasoning.

The results in Table \ref{tab:comp} demonstrate that our method outperforms baselines across all metrics, particularly on 3D VG tasks that require higher spatial reasoning abilities. The results further indicate that QuatRoPE can effectively convey spatial information by providing relative object positions, verifying the correctness of our approaches.

\subsection{Spatial Reasoning Ability Verification}

In the previous experiment, the improved scores across datasets generally indicate that the proposed methods can enhance models' spatial reasoning abilities. To exclusively demonstrate QuatRoPE's effectiveness in enhancing models' spatial reasoning ability, we utilize our ASR benchmark for further evaluation. We conduct zero-shot comparisons between models with and without QuatRoPE. The results are shown in Table \ref{tab:sr_bench}.

The results in the table demonstrate that our model has achieved consistent and substantial gains throughout different experimental settings, directly verifying that the proposed method can enhance models' performance in spatial reasoning. The results further indicate that our method can effectively provide useful relative spatial information to LLMs for solving 3D VL tasks.

\subsection{Ablation Study}

To compare different RoPE settings, including composition approaches (i.e., IGRE and traditional additive approach) and RoPE methods, we perform an ablation study on these factors. Specifically, we utilize baseline models with Llama-3.2-1B-Instruct as the LLM. Then, we apply QuatRoPE via different composition methods, namely, Trans-Additive (where QuatRoPE and language RoPE simultaneously rotate query/key vectors, but with inverse frequencies to lower interference) and IGRE. Results are shown in Table \ref{tab:abl_comp}.
The results demonstrate that among different composition methods, IGRE surpasses the baseline and the model using the Trans-Additive approach under all metrics.
Particularly, IGRE has obtained significant improvements on the 3D VG dataset ScanRefer, where spatial reasoning is the key to locating objects based on spatial relations. The results verify that IGRE can separate QuatRoPE from language RoPE better and minimize interference.

Moreover, we also compare the performance of different positional encoding approaches (i.e., without explicit encoding, directly adding raw $(x,y,z)$ coordinates to feature vectors, M-RoPE \cite{qwen2-vl}, and QuatRoPE) using IGRE.
The results are shown in Table \ref{tab:abl_rope}.
Models with explicit positional encoding outperform baseline models in most cases, suggesting that explicit positional encoding generally improves scene understanding. However, adding raw coordinates introduces absolute positions into the feature vectors and prevents them from being transformed into relative positions during the attention mechanism. Thus, such an approach disrupts models' understanding of scene layouts, especially in models like 3DGraphLLM that rely heavily on input tokens to understand them.
Among RoPE-based approaches, QuatRoPE outperforms M-RoPE across all metrics, demonstrating that encoding coordinates as holistic vectors can convey spatial relations between objects and represent scene layout more effectively.

\begin{table}[t]
    \centering
    \small
    \caption{Verification of advantage in holistic encoding.} \label{tab:resplit}
    \begin{tabular}{cccc}
    \toprule
        $\delta$ & 3DGraphLLM-1B & + QuatRoPE & Gain \\ \midrule
        1 (All) & 93.72 & 94.65 & 0.93 \\
        0.5 & 92.28  & 94.21  & 1.93  \\ 
        0.3 & 91.47  & 94.31  & 2.84  \\ 
        0.2 & 93.21  & 96.30  & 3.09  \\ 
        0.1 & 92.39  & 96.74  & 4.35  \\ 
        0.05 & 84.62  & 92.31  & 7.69  \\ \bottomrule
    \end{tabular}
\end{table}

\begin{figure}[tp]
    \centering
    \includegraphics[width=0.48\textwidth]{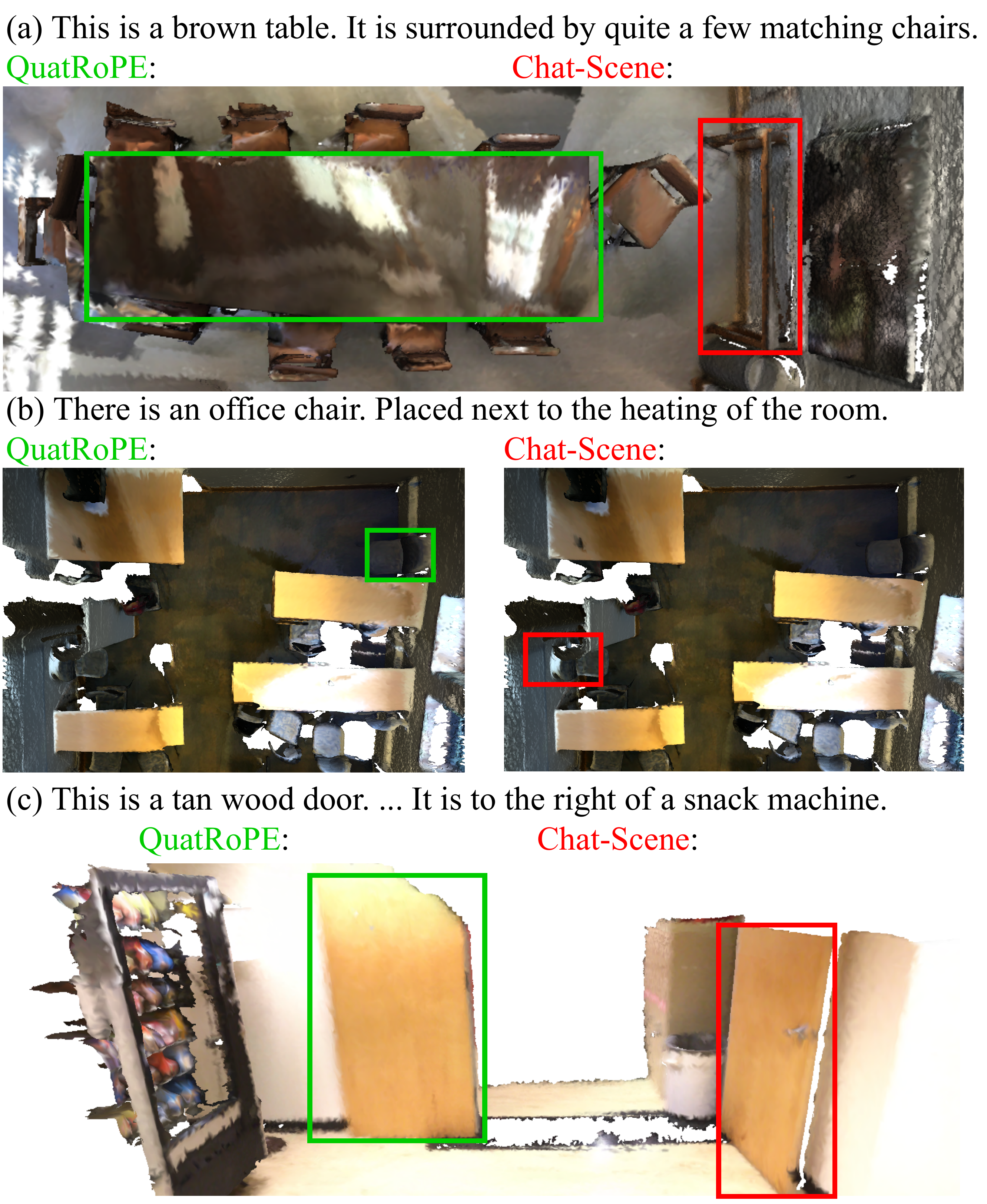}
    \caption{Qualitative results on the ScanRefer dataset. Target objects are correctly grounded by QuatRoPE (\textcolor[HTML]{00cc00}{green}), whereas the baseline Chat-Scene produces incorrect predictions (\textcolor[HTML]{ff0000}{red}).}
    \label{fig:qr}
\end{figure}

\subsection{Verification of Advantage in Holistic Encoding}

In previous RoPE, each axis is treated independently, causing close coordinates on a single axis to falsely appear ``nearby'' and disrupt attention. To address this issue, QuatRoPE encodes positions as holistic vectors.

To verify the effectiveness of such a design, we re-split the ScanRefer \cite{scanrefer} dataset according to the severity of the ``false nearby'' issue.
Since such an issue occurs when the coordinate difference is small on some axis, severity is defined by the aspect ratio $\frac{\min\{\Delta x, \Delta y\}}{\max\{\Delta x, \Delta y\}}<\delta$ for anchor-target object position differences $(\Delta x, \Delta y, \Delta z)$, with lower $\delta$ denoting more severe cases.
As in Tab. \ref{tab:resplit}, QuatRoPE outperforms the 3DGraphLLM baseline in all cases, and its advantage increases with severity, verifying the effectiveness.

\subsection{Qualitative Results}

Finally, we visualize several cases in ScanRefer to demonstrate the effectiveness of our approach. The qualitative results are illustrated in Fig. \ref{fig:qr}.

In these cases, as relative positions between objects are well represented, models can better align spatial information with descriptions like ``surrounded by'' and ``next to''.
Notably, in Case (c), while both doors are to the machine's right, the correct one is closer (which can be explained by humans' preference for the ``Maxim of Relation'' \cite{logic_and_conv} in linguistics). Such a case also indicates that, as QuatRoPE correctly increases attention scores for proximate objects, models can align with human implication better, enabling models to understand and predict human-like spatial reasoning patterns across multimodal tasks.
\section{Conclusion}

In this paper, we propose QuatRoPE, a positional embedding that encodes objects' positions to tokens and leverages the attention mechanism to transform absolute positions into objects' pairwise spatial relations. To minimize QuatRoPE's interference with language RoPE, we further propose IGRE for separating dimensions for RoPEs and constraining QuatRoPE's effect to object-related tokens. Extensive experiments demonstrate the effectiveness of QuatRoPE and IGRE. Moreover, through our ASR benchmark, we verify that our method can achieve large gains in spatial reasoning across various baselines, offering a solution for enhancing the spatial reasoning ability of 3D LLMs.

\newpage

\noindent\textbf{Acknowledgements.} This work was supported by the grants from the Beijing Natural Science Foundation 4252040, the Beijing Nova Program, and the National Natural Science Foundation of China 62372014.

{
    \small
    \bibliographystyle{ieeenat_fullname}
    \bibliography{main}
}

\newpage

\appendix
\section{Mathematical Derivation for QuatRoPE}

In this section, we provide a detailed mathematical derivation for QuatRoPE.

Let $\vec m$ and $\vec n$ be the absolute 3D coordinates of the objects corresponding to query vector $\vec q$ and key vector $\vec k$, and $f(\vec x, \vec p)$ be the function for rotating the query or key vector $\vec x$ with a corresponding 3D position $\vec p$. Since the attention score should only relate to the relative position (i.e., $\vec m-\vec n$), the rotation function $f$ should satisfy:

\begin{equation} \label{eq:supp_quat_goal}
    \left<f(\vec q, \vec m), f(\vec k, \vec n)\right> = g(\vec q, \vec k, \vec m-\vec n)
\end{equation}

In QuatRoPE, we embed the coordinates as a holistic vector by applying quaternion rotations to query and key vectors. Formally, the rotation function can be expressed as:

\begin{equation} \label{eq:supp_quat_repr}
    \begin{cases}
        f(\vec q, \vec m)=Q(\vec m)~\vec q~Q^{-1}(\vec m)\\
        Q(\vec m)=Q_z(m_z)~Q_y(m_y)~Q_x(m_x)\\
        Q_x(m_x)=\cos\left[\theta_x(m_x)/2\right]+\hat i\sin\left[\theta_x(m_x)/2\right]\\
        Q_y(m_y)=\cos\left[\theta_y(m_y)/2\right]+\hat j\sin\left[\theta_y(m_y)/2\right]\\
        Q_z(m_z)=\cos\left[\theta_z(m_z)/2\right]+\hat k\sin\left[\theta_z(m_z)/2\right]
    \end{cases}
\end{equation}

\noindent where $Q$'s are rotation matrices and $\theta$'s are unary functions.

Through Equation (\ref{eq:supp_quat_repr}), we transform the requirement in QuatRoPE (i.e., converting absolute coordinates to relative positions via dot products) into deriving $\theta$'s that satisfy Equation (\ref{eq:supp_quat_goal}).
To solve the equation, we transform the dot product into the real part of the product of the rotation functions to yield a form with multiplication between the rotation matrices (i.e., $Q^{-1}(\vec m)$ and $Q(\vec n)$).

\begin{equation} \label{eq:supp_quat_exp}
    \begin{aligned}
        &\left<f(\vec q, \vec m), f(\vec k, \vec n)\right>\\
        =&\Re[f(\vec q, \vec m)f^*(\vec k, \vec n)]\\
        =&\Re[Q(\vec m)~\vec q~Q^{-1}(\vec m)~\overline{Q(\vec n)~\vec k~Q^{-1}(\vec n)}]\\
        =&\Re[Q(\vec m)~\vec q~Q^{-1}(\vec m)~Q(\vec n)~\vec k^*~Q^{-1}(\vec n)]
    \end{aligned}
\end{equation}

\noindent where $\vec k^*$ denotes the conjugate of quaternion $\vec k$, and $\Re$ denotes the real part of the quaternion. To pair every $Q(\vec m)$ with $Q(\vec n)$, according to the real-part invariance of quaternion rotation, after left multiplying $Q^{-1}(\vec m)$ and right multiplying $Q(\vec m)$, Equation (\ref{eq:supp_quat_exp}) yields:

\begin{equation}
    \begin{aligned}
        &\left<f(\vec q, \vec m), f(\vec k, \vec n)\right>\\
        =&\Re[Q(\vec m)~\vec q~Q^{-1}(\vec m)~Q(\vec n)~\vec k^*~Q^{-1}(\vec n)~Q(\vec m)~Q^{-1}(\vec m)]\\
        =&\Re[\vec q~Q^{-1}(\vec m)~Q(\vec n)~\vec k^*~Q^{-1}(\vec n)~Q(\vec m)]
    \end{aligned}
\end{equation}

According to Equation (\ref{eq:supp_quat_goal}), $\left<f(\vec q, \vec m), f(\vec k, \vec n)\right>$ should only relate to $\vec m-\vec n$, the following equation should hold:

\begin{equation}
    \Re[\vec q~Q^{-1}(\vec m)~Q(\vec n)~\vec k^*~Q^{-1}(\vec n)~Q(\vec m)]=g(\vec q, \vec k, \vec m-\vec n)
\end{equation}

Thus,

\begin{equation} \label{eq:supp_abs_funct}
    Q(\vec m-\vec n)=Q^{-1}(\vec n)~Q(\vec m)
\end{equation}

\noindent i.e.,

\begin{equation} \label{eq:supp_exp_q}
    \begin{aligned}
        &Q_z(m_z-n_z)~Q_y(m_y-n_y)~Q_x(m_x-n_x)\\
        =&Q_x^{-1}(n_x)~Q_y^{-1}(n_y)~Q_z^{-1}(n_z)~Q_z(m_z)~Q_y(m_y)~Q_x(m_x)
    \end{aligned}
\end{equation}

When $\vec m=\vec n=\vec 0$, we have $Q(\vec 0)~Q(\vec 0)=Q(\vec 0)$. Thus, $Q(\vec 0)=1$. According to Equation (\ref{eq:supp_quat_repr}),

\begin{equation} \label{eq:supp_exp_0}
    \begin{aligned}
        1=&Q(\vec 0)\\
        =&Q_z(0)~Q_y(0)~Q_x(0)\\
        =&\left[\cos\left(\dfrac{\theta_z(0)}2\right)+\hat k\sin\left(\dfrac{\theta_z(0)}2\right)\right]\\
        &\left[\cos\left(\dfrac{\theta_y(0)}2\right)+\hat j\sin\left(\dfrac{\theta_y(0)}2\right)\right]\\
        &\left[\cos\left(\dfrac{\theta_x(0)}2\right)+\hat i\sin\left(\dfrac{\theta_x(0)}2\right)\right]
    \end{aligned}
\end{equation}

Consider the real part of the equation above, we have:

\begin{equation}
    \begin{aligned}
        1=&\Re\left\{\left[\cos\left(\dfrac{\theta_z(0)}2\right)+\hat k\sin\left(\dfrac{\theta_z(0)}2\right)\right]\right.\\
        &\left[\cos\left(\dfrac{\theta_y(0)}2\right)+\hat j\sin\left(\dfrac{\theta_y(0)}2\right)\right]\\
        &\left.\left[\cos\left(\dfrac{\theta_x(0)}2\right)+\hat i\sin\left(\dfrac{\theta_x(0)}2\right)\right]\right\}\\
        =&\cos\left(\dfrac{\theta_z(0)}2\right)\cos\left(\dfrac{\theta_y(0)}2\right)\cos\left(\dfrac{\theta_x(0)}2\right)\\
        &+\hat k\hat j\hat i\sin\left(\dfrac{\theta_z(0)}2\right)\sin\left(\dfrac{\theta_y(0)}2\right)\sin\left(\dfrac{\theta_x(0)}2\right)\\
        =&\cos\left(\dfrac{\theta_x(0)}2\right)\cos\left(\dfrac{\theta_y(0)}2\right)\cos\left(\dfrac{\theta_z(0)}2\right)\\
        &+\sin\left(\dfrac{\theta_x(0)}2\right)\sin\left(\dfrac{\theta_y(0)}2\right)\sin\left(\dfrac{\theta_z(0)}2\right)
    \end{aligned}
\end{equation}

Also, since the imaginary part of Equation (\ref{eq:supp_exp_0}) is $0$, either all cosines or all sines are equal to $0$. Therefore

\begin{equation} \label{eq:supp_cos_case_1}
    \cos\left(\dfrac{\theta_x(0)}2\right)=\cos\left(\dfrac{\theta_y(0)}2\right)=\cos\left(\dfrac{\theta_z(0)}2\right)=1
\end{equation}

\noindent or

\begin{equation} \label{eq:supp_sin_case_1}
    \sin\left(\dfrac{\theta_x(0)}2\right)=\sin\left(\dfrac{\theta_y(0)}2\right)=\sin\left(\dfrac{\theta_z(0)}2\right)=1
\end{equation}

Consider the first solution, let $m_x=m_y=n_x=n_y=0$, Equation (\ref{eq:supp_exp_q}) yields:

\begin{table*}[!ht]
    \centering
    \caption{Comparison between fixed and learnable base vectors for rotation.} \label{tab:supp_comp}
    \begin{tabular}{cccccccc}
    \toprule
        \multirow{2}{*}{Model} & \multirow{2}{*}{Base Vector} & \multicolumn{2}{c}{ScanRefer} & SQA3D & \multicolumn{2}{c}{Multi3dRef} \\ \cmidrule(lr){3-4} \cmidrule(lr){5-5} \cmidrule(lr){6-7}
        ~ & ~ & Acc @ 0.25 & Acc @ 0.5 & EM @ 1 & F1 @ 0.25 & F1 @ 0.5 \\ \midrule
        \multirow{2}{*}{Chat-Scene \cite{chat-scene}} & Fixed & 55.44 & 55.00 & 53.14 & 58.09 & 57.72 \\ 
        ~ & Learnable & 54.47 & 54.14 & 52.84 & 57.96 & 57.74 \\ \midrule
        \multirow{2}{*}{3DGraphLLM \cite{3dgraphllm}} & Fixed & 58.30 & 58.15 & 53.20 & 60.70 & 60.52 \\ 
        ~ & Learnable & 56.89 & 56.64 & 52.68 & 60.67 & 60.51 \\ \bottomrule
    \end{tabular}
\end{table*}
\begin{table*}[!ht]
    \centering
    \caption{Comparison between different frequencies.} \label{tab:supp_freq}
    \begin{tabular}{cccccc}
    \toprule
        \multirow{2}{*}{Frequency} & \multicolumn{2}{c}{ScanRefer} & SQA3D & \multicolumn{2}{c}{Multi3dRef} \\ \cmidrule(lr){2-3} \cmidrule(lr){4-4} \cmidrule(lr){5-6}
            ~ & Acc @ 0.25 & Acc @ 0.5 & EM @ 1 & F1 @ 0.25 & F1 @ 0.5 \\ \midrule
            0.3 (Default) & \textbf{58.30}  & \textbf{58.15}  & \textbf{60.70}  & \textbf{60.52}  & \textbf{53.20}  \\ 
            0.1 (Small) & 54.55  & 54.39  & 58.02  & 57.90  & 51.99  \\ 
            1.0 (Large) & 53.41  & 53.14  & 56.28  & 55.99  & 52.18  \\ \bottomrule
    \end{tabular}
\end{table*}

\begin{equation}
    \begin{aligned}
        &Q_z(m_z-n_z)~Q_y(0-0)~Q_x(0-0)\\
        =&Q_x^{-1}(0)~Q_y^{-1}(0)~Q_z^{-1}(n_z)~Q_z(m_z)~Q_y(0)~Q_x(0)
    \end{aligned}
\end{equation}

\noindent i.e.,

\begin{equation} \label{eq:supp_imp_eq_z}
    Q_z(m_z-n_z)=Q_z^{-1}(n_z)~Q_z(m_z)
\end{equation}

For Equation (\ref{eq:supp_imp_eq_z}), the left-hand side

\begin{equation} \label{eq:supp_lhs}
    \begin{aligned}
        &Q_z(m_z-n_z)\\
        =&\cos\left(\dfrac{\theta_z(m_z-n_z)}2\right)+\sin\left(\dfrac{\theta_z(m_z-n_z)}2\right)\hat k
    \end{aligned}
\end{equation}

\noindent while the right-hand side

\begin{equation} \label{eq:supp_rhs}
    \begin{aligned}
        &Q_z^{-1}(n_z)~Q_z(m_z)\\
        =&\left[\cos\left(\theta_z(n_z)/2\right)-\hat k\sin\left(\theta_z(n_z)/2\right)\right]\\
        &\left[\cos\left(\theta_z(m_z)/2\right)+\hat k\sin\left(\theta_z(m_z)/2\right)\right]\\
        =&\left[\cos\left(\theta_z(n_z)/2\right)\cos\left(\theta_z(m_z)/2\right)\right.\\
        &\left.+\sin\left(\theta_z(n_z)/2\right)\sin\left(\theta_z(m_z)/2\right)\right]\\
        &+\left[\cos\left(\theta_z(n_z)/2\right)\sin\left(\theta_z(m_z)/2\right)\right.\\
        &-\left.\sin\left(\theta_z(n_z)/2\right)\cos\left(\theta_z(m_z)/2\right)\right]\hat k \\
        =&\cos\left(\dfrac{\theta_z(m_z)-\theta_z(n_z)}2\right)+\sin\left(\dfrac{\theta_z(m_z)-\theta_z(n_z)}2\right)\hat k
    \end{aligned}
\end{equation}

By Equation (\ref{eq:supp_lhs}) and Equation (\ref{eq:supp_rhs}), we have

\begin{equation} \label{eq:supp_theta_z_eq}
    \theta_z(m_z)-\theta_z(n_z)=\theta_z(m_z-n_z)
\end{equation}

When $m_z=n_z$, Equation (\ref{eq:supp_theta_z_eq}) yields:

\begin{equation}
    \begin{aligned}
        \theta_z(0)&=\theta_z(m_z-n_z)\\
        &=\theta_z(m_z)-\theta_z(n_z)\\
        &=0
    \end{aligned}
\end{equation}

Then, for any $t\in\mathbb Z$, we have

\begin{equation}
    \begin{aligned}
        \theta_z(t)=&\theta_z(t-1)+\theta_z(1)\\
        =&\theta_z(t-2)+\theta_z(1)+\theta_z(1)\\
        =&\cdots\\
        =&\theta_z(0)+t\theta_z(1)\\
        =&t\theta_z(1)
    \end{aligned}
\end{equation}

Moreover, for any $t, p\in\mathbb Z$ and $(t, p)=1$ (i.e., $\dfrac tp\in\mathbb Q$), we have

\begin{equation}
    \begin{aligned}
        \theta_z(t)=&\theta_z\left(\dfrac{t(p-1)}p\right)+\theta_z\left(\dfrac tp\right)\\
        =&\theta_z\left(\dfrac{t(p-2)}p\right)+\theta_z\left(\dfrac tp\right)+\theta_z\left(\dfrac tp\right)\\
        =&\cdots\\
        =&p\theta_z\left(\dfrac tp\right)
    \end{aligned}
\end{equation}

\noindent and hence

\begin{equation}
    \theta_z\left(\dfrac tp\right)=\dfrac 1p\theta_z(t)=\dfrac tp\theta_z(1)
\end{equation}

Also, since the embedding should be continuous with respect to the position, $\theta_z$ should be continuous, and the solution to $\theta_z$ is

\begin{equation}
    \theta_z(z)=z\theta_z(1), z\in\mathbb R
\end{equation}

Let $n_z=m_z=0$, according to Equation (\ref{eq:supp_exp_q}), we have

\begin{equation} \label{eq:supp_exp_q2}
    \begin{aligned}
        &Q_y(m_y-n_y)~Q_x(m_x-n_x)\\
        =&Q_x^{-1}(n_x)~Q_y^{-1}(n_y)~Q_y(m_y)~Q_x(m_x)
    \end{aligned}
\end{equation}

Similarly, the above equation yields

\begin{equation}
    \theta_y(y)=y\theta_y(1), y\in\mathbb R
\end{equation}

Again, let $n_y=m_y=n_z=m_z=0$, Equation (\ref{eq:supp_exp_q2}) yields

\begin{equation}
    Q_x(m_x-n_x)=Q_x^{-1}(n_x)~Q_x(m_x)
\end{equation}

\noindent and thus

\begin{equation}
    \theta_x(x)=x\theta_x(1), x\in\mathbb R
\end{equation}

In conclusion, an approximate solution for QuatRoPE is:

\begin{equation}
    \begin{cases}
        f(\vec q, \vec m)=Q(\vec m)~\vec q~Q^{-1}(\vec m)\\
        Q(\vec m)=Q_z(m_z)~Q_y(m_y)~Q_x(m_x)\\
        Q_x(m_x)=\cos\left[\dfrac{m_x\theta_x(1)}2\right]+\hat i\sin\left[\dfrac{m_x\theta_x(1)}2\right]\\
        Q_y(m_y)=\cos\left[\dfrac{m_y\theta_y(1)}2\right]+\hat j\sin\left[\dfrac{m_y\theta_y(1)}2\right]\\
        Q_z(m_z)=\cos\left[\dfrac{m_z\theta_z(1)}2\right]+\hat k\sin\left[\dfrac{m_z\theta_z(1)}2\right]
    \end{cases}
\end{equation}

\noindent where $\theta_x(1)$, $\theta_y(1)$, and $\theta_z(1)$ are frequencies for quaternion rotations.
According to Equation (\ref{eq:supp_quat_goal}), as we perform rotation by $\vec q:=f(\vec q,\vec m)$ and $\vec k:=f(\vec k, \vec n)$ before each attention layer, the attention scores between object-related tokens reflect their relative positions. By such an approach, QuatRoPE can effectively convey relative positional information for LLMs to perform spatial reasoning.
\section{Experimental Settings}
\begin{table*}[!ht]
    \centering
    \caption{Qualitative Results} \label{tab:qr1}
    \vskip 0.05in
    \hrule
    \vskip 0.10in
    \begin{minipage}{0.28\textwidth}
        \centering
        \textbf{Description}
    \end{minipage}
    \begin{minipage}{0.35\textwidth}
        \centering
        \textbf{Chat-Scene \cite{chat-scene}}
    \end{minipage}
    \begin{minipage}{0.35\textwidth}
        \centering
        \textbf{QuatRoPE (Ours)}
    \end{minipage}\hfill
    
    \vskip 0.10in
    \hrule
    \vskip 0.10in
    
    \begin{minipage}{0.28\textwidth}
        This is a brown chair. It is turned toward the end of the table.
    \end{minipage}
    \begin{minipage}{0.35\textwidth}
        \centering
        \includegraphics[width=0.98\linewidth]{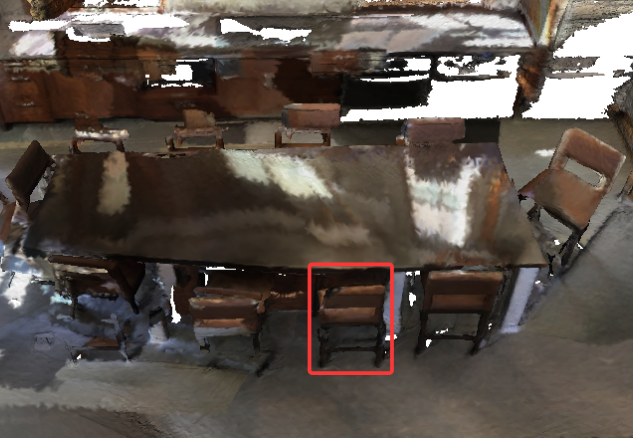}
    \end{minipage}
    \begin{minipage}{0.35\textwidth}
        \centering
        \includegraphics[width=0.98\linewidth]{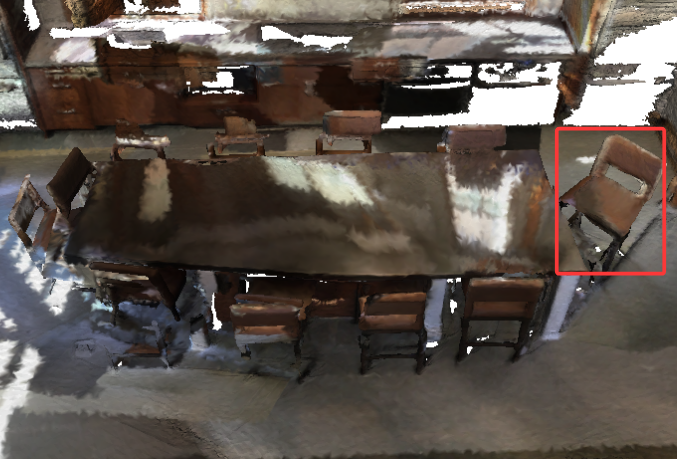}
    \end{minipage}\hfill
    
    \vskip 0.10in
    \hrule
    \vskip 0.10in
    
    \begin{minipage}{0.28\textwidth}
        Box-shaped footstool with a tarnished red color. There are 6 footstools stacked, 3 on the bottom row and 3 on the top. This is located on the bottom row in the middle.
    \end{minipage}
    \begin{minipage}{0.35\textwidth}
        \centering
        \includegraphics[width=0.98\linewidth]{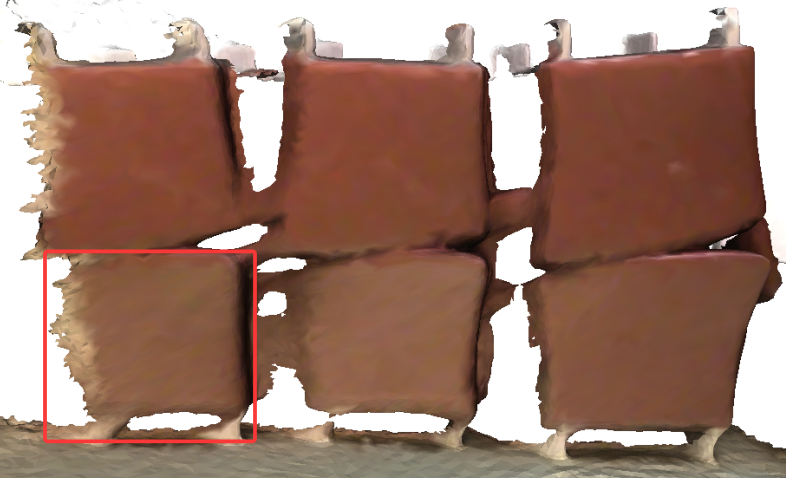}
    \end{minipage}
    \begin{minipage}{0.35\textwidth}
        \centering
        \includegraphics[width=0.98\linewidth]{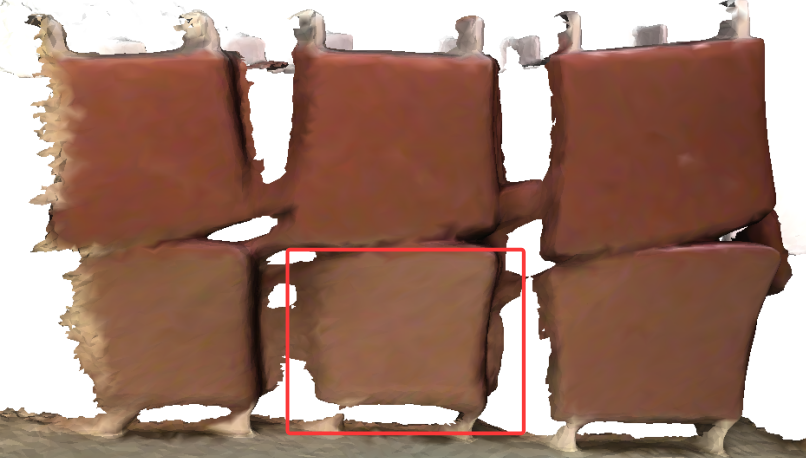}
    \end{minipage}\hfill
    
    \vskip 0.10in
    \hrule
    \vskip 0.10in
    
    \begin{minipage}{0.28\textwidth}
        A blue towel that is hanging on the glass shower door. The towel is in the middle of the three towels hanging on the shower handle.
    \end{minipage}
    \begin{minipage}{0.35\textwidth}
        \centering
        \includegraphics[width=0.98\linewidth]{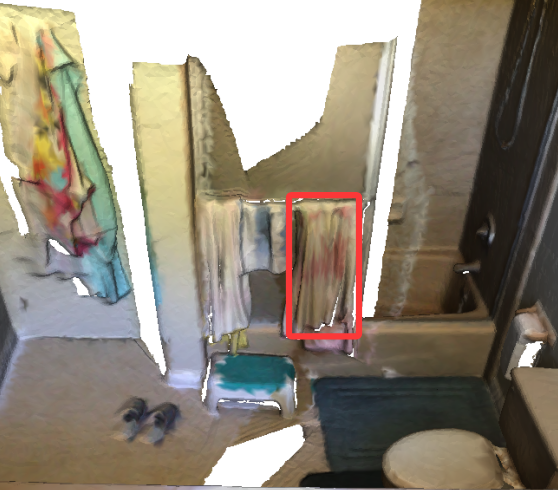}
    \end{minipage}
    \begin{minipage}{0.35\textwidth}
        \centering
        \includegraphics[width=0.98\linewidth]{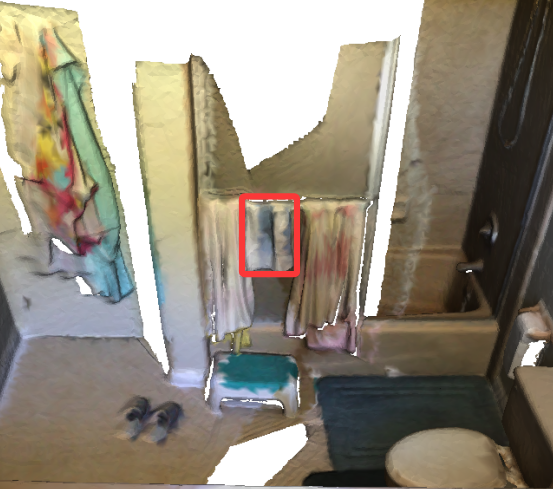}
    \end{minipage}\hfill
    
    \vskip 0.10in
    \hrule
    \vskip 0.10in
    
    \begin{minipage}{0.28\textwidth}
        This is a black office chair. It is facing the desk corner.
    \end{minipage}
    \begin{minipage}{0.35\textwidth}
        \centering
        \includegraphics[width=0.98\linewidth]{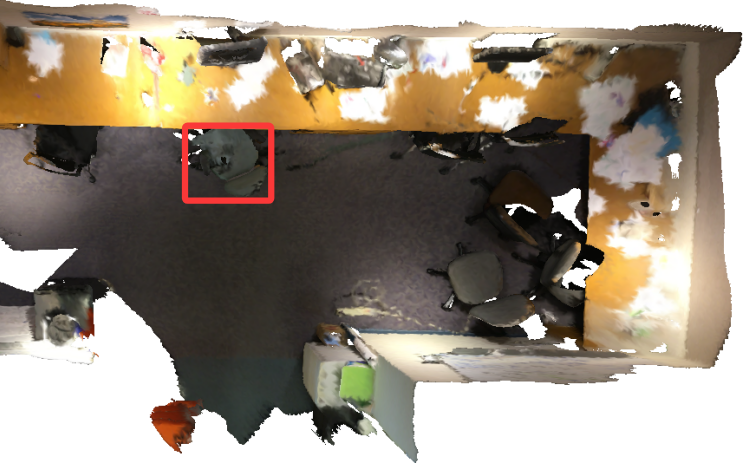}
    \end{minipage}
    \begin{minipage}{0.35\textwidth}
        \centering
        \includegraphics[width=0.98\linewidth]{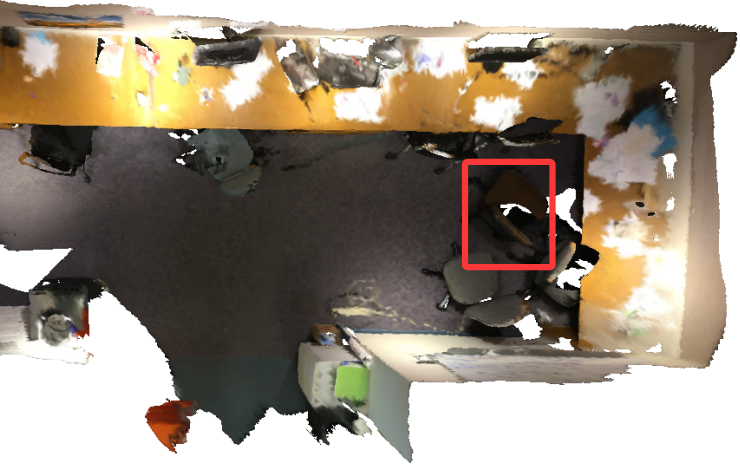}
    \end{minipage}\hfill
    
    \vskip 0.10in
    \hrule
\end{table*}

\subsection{Base Vector for Rotation}

In IGRE, the quaternion rotation of QuatRoPE is applied to the base vector to obtain the positional embedding. In this section, we compare the performance between using $(1, 0, 0)$ as a fixed base vector and the strategy of using a learnable base vector. Then we train and evaluate these approaches on Chat-Scene-1B \cite{chat-scene} and 3DGraphLLM-1B \cite{3dgraphllm}, and the results are shown in Table \ref{tab:supp_comp}.

The results indicate that learnable base vectors do not achieve better results. Such outcomes may result from the difficulty of learning base vectors, as these vectors have a significant impact on subsequent layers. Therefore, in our model, we set the base vector as $(1, 0, 0)$, which is also more computationally efficient.

\subsection{Choice of Rotation Frequency}

In the experiments, rotation frequency is set to 0.3 (untuned, consistent across all datasets) to avoid two issues shown in Tab. \ref{tab:supp_freq}:
(a) Small frequencies lead to small rotation angles, weakening feature vector influence and hindering learning.
(b) Large frequencies cause the ``wrapping'' problem—large coordinate differences may produce similar rotation angles, misleading the model with incorrect scene layouts.

Given the maximum coordinate difference of 10, frequency is set to $\frac\pi{10}\approx0.3$, ensuring all rotations lie in the same semi-circle and larger coordinate differences correspond to larger angle differences.

Additionally, the error introduced by the non‑commutativity of the Euler angle decomposition sequence is proportional to the square of the frequency. Thus, selecting a small frequency (e.g., 0.3) also makes QuatRoPE closer to the requirement of Equation (\ref{eq:supp_quat_goal}).
\section{Qualitative Results}

\begin{table*}[!ht]
    \centering
    \caption{Qualitative Results (Continued)} \label{tab:qr2}
    \vskip 0.05in
    \hrule
    \vskip 0.05in
    \begin{minipage}{0.28\textwidth}
        \centering
        \textbf{Description}
    \end{minipage}
    \begin{minipage}{0.35\textwidth}
        \centering
        \textbf{Chat-Scene \cite{chat-scene}}
    \end{minipage}
    \begin{minipage}{0.35\textwidth}
        \centering
        \textbf{QuatRoPE (Ours)}
    \end{minipage}\hfill
    
    \vskip 0.05in
    \hrule
    \vskip 0.10in
    
    \begin{minipage}{0.28\textwidth}
        It is a brown chair with armrests and four legs. It is directly under a blackboard.
    \end{minipage}
    \begin{minipage}{0.35\textwidth}
        \centering
        \includegraphics[width=0.98\linewidth]{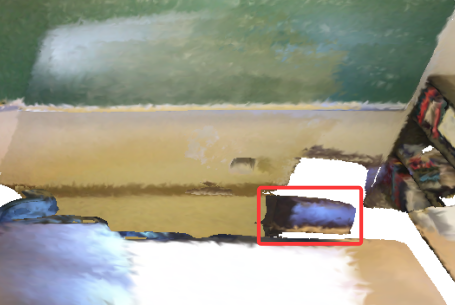}
    \end{minipage}
    \begin{minipage}{0.35\textwidth}
        \centering
        \includegraphics[width=0.98\linewidth]{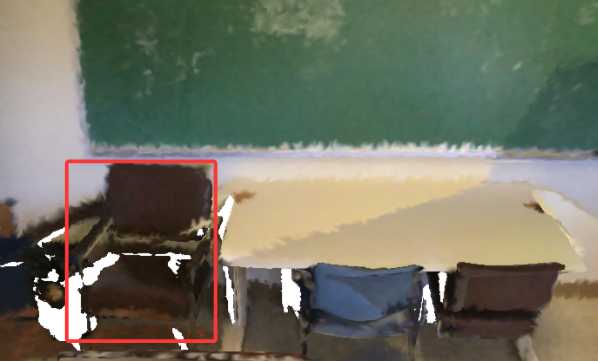}
    \end{minipage}\hfill
    
    \vskip 0.10in
    \hrule
    \vskip 0.10in
    
    \begin{minipage}{0.28\textwidth}
        Case 1: This door appears to be the front door to the apartment. If you walk through the apartment and past the bathroom, you will encounter this door. The door is black and has a small window.\\
        Case 2: The door is rectangular in shape and has a small window on the upper portion. The door is located to the right of the bath area.
        Chat-Scene fails under both cases.
    \end{minipage}
    \begin{minipage}{0.35\textwidth}
        \centering
        \includegraphics[width=0.98\linewidth]{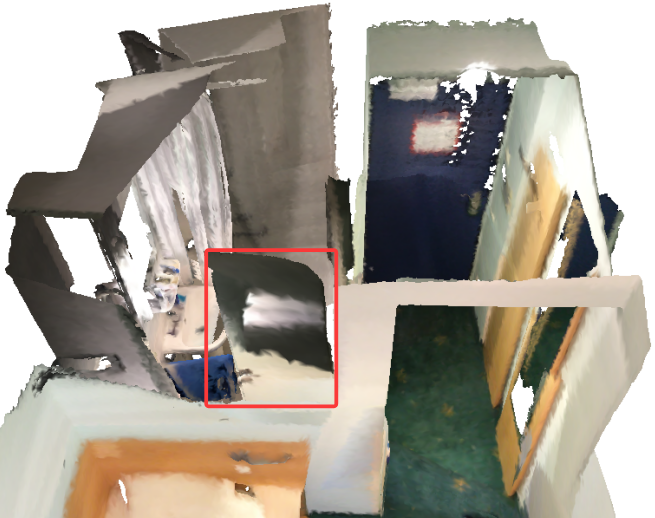}
    \end{minipage}
    \begin{minipage}{0.35\textwidth}
        \centering
        \includegraphics[width=0.98\linewidth]{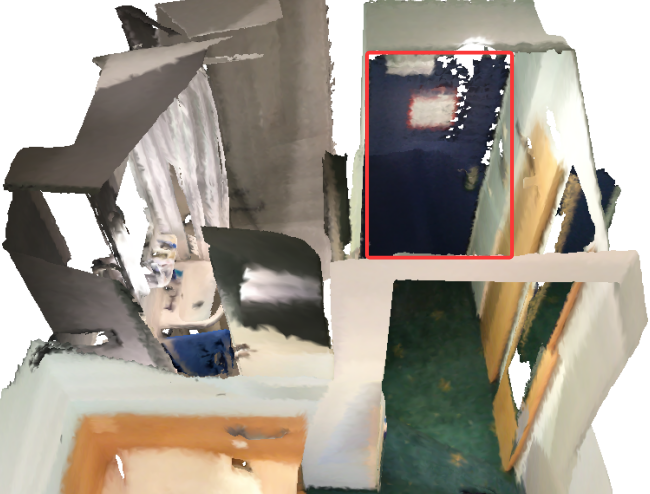}
    \end{minipage}\hfill
    
    \vskip 0.10in
    \hrule
    \vskip 0.10in
    
    % \begin{minipage}{0.28\textwidth}
    %     The floral-printed chair is in front of the glass double doors. The floral-printed chair is to the left of the grey desk.
    % \end{minipage}
    % \begin{minipage}{0.35\textwidth}
    %     \centering
    %     \includegraphics[width=0.98\linewidth]{supp_figs/err_50_1.png}
    % \end{minipage}
    % \begin{minipage}{0.35\textwidth}
    %     \centering
    %     \includegraphics[width=0.98\linewidth]{supp_figs/gt_50_1.png}
    % \end{minipage}\hfill
    
    % \vskip 0.10in
    % \hrule
    % \vskip 0.10in
    
    % \begin{minipage}{0.28\textwidth}
    %     This is a long bookshelf. It is next to some brown armchairs.
    % \end{minipage}
    % \begin{minipage}{0.35\textwidth}
    %     \centering
    %     \includegraphics[width=0.98\linewidth]{supp_figs/err_64_1.png}
    % \end{minipage}
    % \begin{minipage}{0.35\textwidth}
    %     \centering
    %     \includegraphics[width=0.98\linewidth]{supp_figs/gt_64_1.png}
    % \end{minipage}\hfill

    \begin{minipage}{0.28\textwidth}
        The small rounded table. The table is next to the couch end.
    \end{minipage}
    \begin{minipage}{0.35\textwidth}
        \centering
        \includegraphics[width=0.98\linewidth]{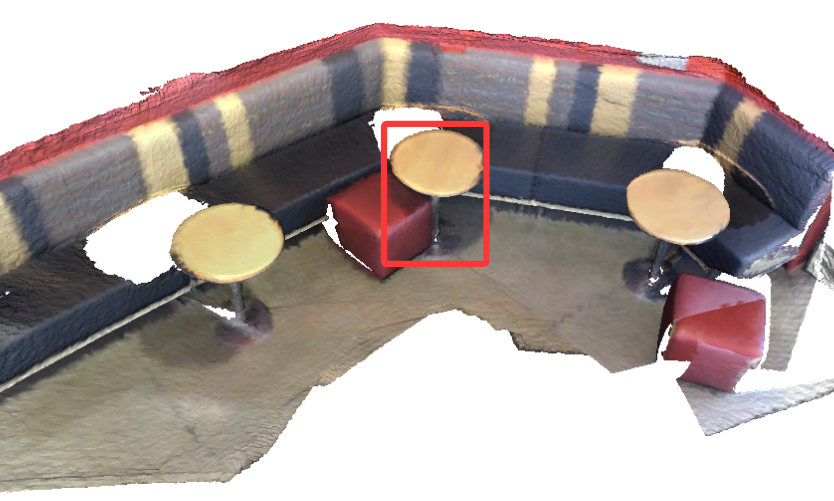}
    \end{minipage}
    \begin{minipage}{0.35\textwidth}
        \centering
        \includegraphics[width=0.98\linewidth]{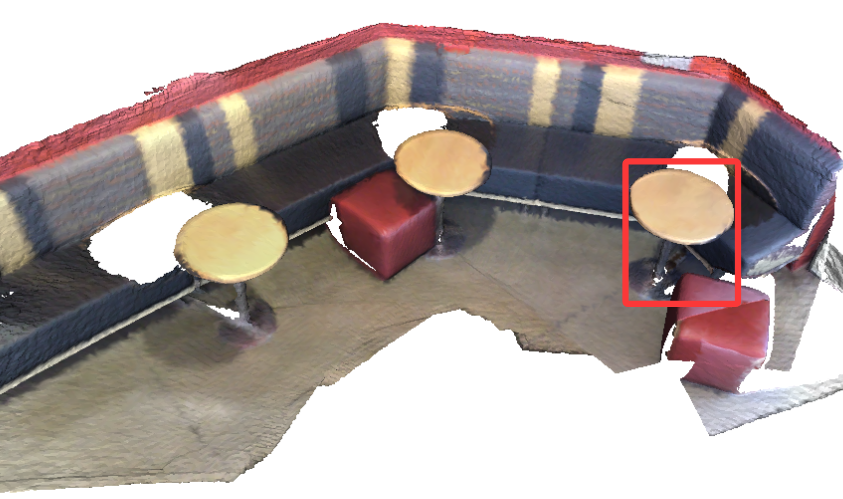}
    \end{minipage}\hfill
    
    \vskip 0.10in
    \hrule
    \vskip 0.10in
    
    \begin{minipage}{0.28\textwidth}
        It is a tall gray trash can. The trash can is under the left side of the counter, to the left of the door when you enter.
    \end{minipage}
    \begin{minipage}{0.35\textwidth}
        \centering
        \includegraphics[width=0.98\linewidth]{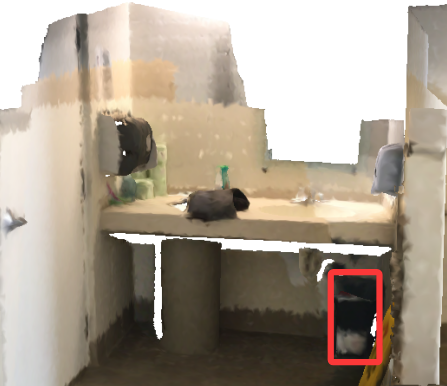}
    \end{minipage}
    \begin{minipage}{0.35\textwidth}
        \centering
        \includegraphics[width=0.98\linewidth]{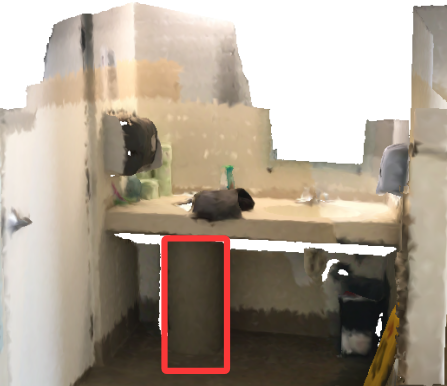}
    \end{minipage}\hfill
    
    \vskip 0.10in
    \hrule
\end{table*}

\begin{table*}[!ht]
    \centering
    \caption{Qualitative Results (Continued)} \label{tab:qr3}
    \vskip 0.05in
    \hrule
    \vskip 0.05in
    \begin{minipage}{0.28\textwidth}
        \centering
        \textbf{Description}
    \end{minipage}
    \begin{minipage}{0.35\textwidth}
        \centering
        \textbf{Chat-Scene \cite{chat-scene}}
    \end{minipage}
    \begin{minipage}{0.35\textwidth}
        \centering
        \textbf{QuatRoPE (Ours)}
    \end{minipage}\hfill
    
    \vskip 0.05in
    \hrule
    \vskip 0.15in
    
    % \begin{minipage}{0.28\textwidth}
    %     The small rounded table. The table is next to the couch end.
    % \end{minipage}
    % \begin{minipage}{0.35\textwidth}
    %     \centering
    %     \includegraphics[width=0.98\linewidth]{supp_figs/err_81_1.png}
    % \end{minipage}
    % \begin{minipage}{0.35\textwidth}
    %     \centering
    %     \includegraphics[width=0.98\linewidth]{supp_figs/gt_81_1.png}
    % \end{minipage}\hfill
    
    % \vskip 0.10in
    % \hrule
    % \vskip 0.10in
    
    % \begin{minipage}{0.28\textwidth}
    %     It is a tall gray trash can. The trash can is under the left side of the counter, to the left of the door when you enter.
    % \end{minipage}
    % \begin{minipage}{0.35\textwidth}
    %     \centering
    %     \includegraphics[width=0.98\linewidth]{supp_figs/err_84_1.png}
    % \end{minipage}
    % \begin{minipage}{0.35\textwidth}
    %     \centering
    %     \includegraphics[width=0.98\linewidth]{supp_figs/gt_84_1.png}
    % \end{minipage}\hfill
    
    % \vskip 0.10in
    % \hrule
    % \vskip 0.10in
    
    % \begin{minipage}{0.28\textwidth}
    %     This is a tan paper towel dispenser. It's a mostly square shape and is to the right of both sinks.
    % \end{minipage}
    % \begin{minipage}{0.35\textwidth}
    %     \centering
    %     \includegraphics[width=0.98\linewidth]{supp_figs/err_86_1.png}
    % \end{minipage}
    % \begin{minipage}{0.35\textwidth}
    %     \centering
    %     \includegraphics[width=0.98\linewidth]{supp_figs/gt_86_1.png}
    % \end{minipage}\hfill
    
    % \vskip 0.10in
    % \hrule
    % \vskip 0.10in
    
    \begin{minipage}{0.28\textwidth}
        Stand in front of the free-standing board in the room.  Looking down the side of the table closest to you, it is the second chair down the row.
    \end{minipage}
    \begin{minipage}{0.35\textwidth}
        \centering
        \includegraphics[width=0.98\linewidth]{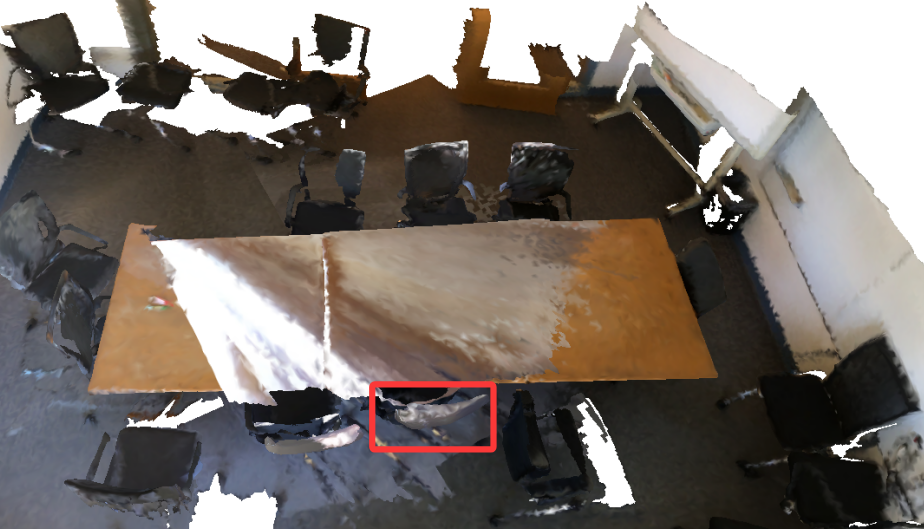}
    \end{minipage}
    \begin{minipage}{0.35\textwidth}
        \centering
        \includegraphics[width=0.98\linewidth]{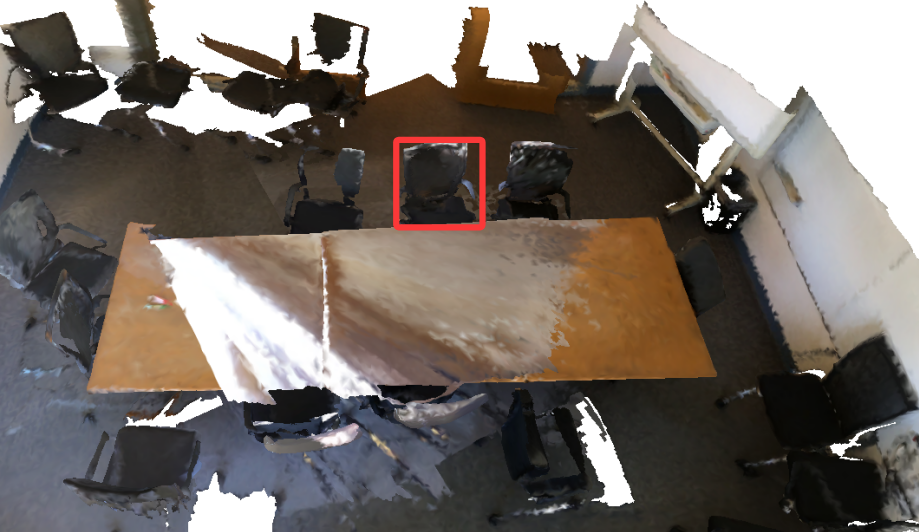}
    \end{minipage}\hfill
    
    \vskip 0.15in
    \hrule
    \vskip 0.15in
    
    \begin{minipage}{0.28\textwidth}
        Case 1: The monitor is next to the leftmost window. The monitor is black and rectangular.\\
        Case 2: The monitor is on the silver table. The monitor is the closest to the window.
    \end{minipage}
    \begin{minipage}{0.35\textwidth}
        \centering
        \includegraphics[width=0.98\linewidth]{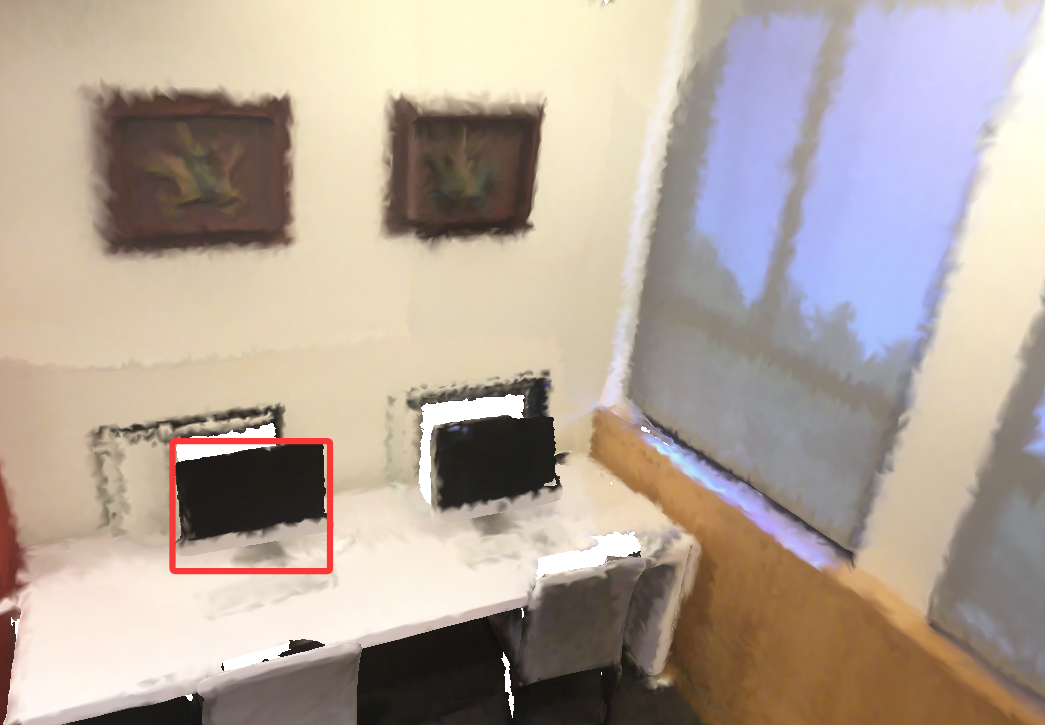}
    \end{minipage}
    \begin{minipage}{0.35\textwidth}
        \centering
        \includegraphics[width=0.98\linewidth]{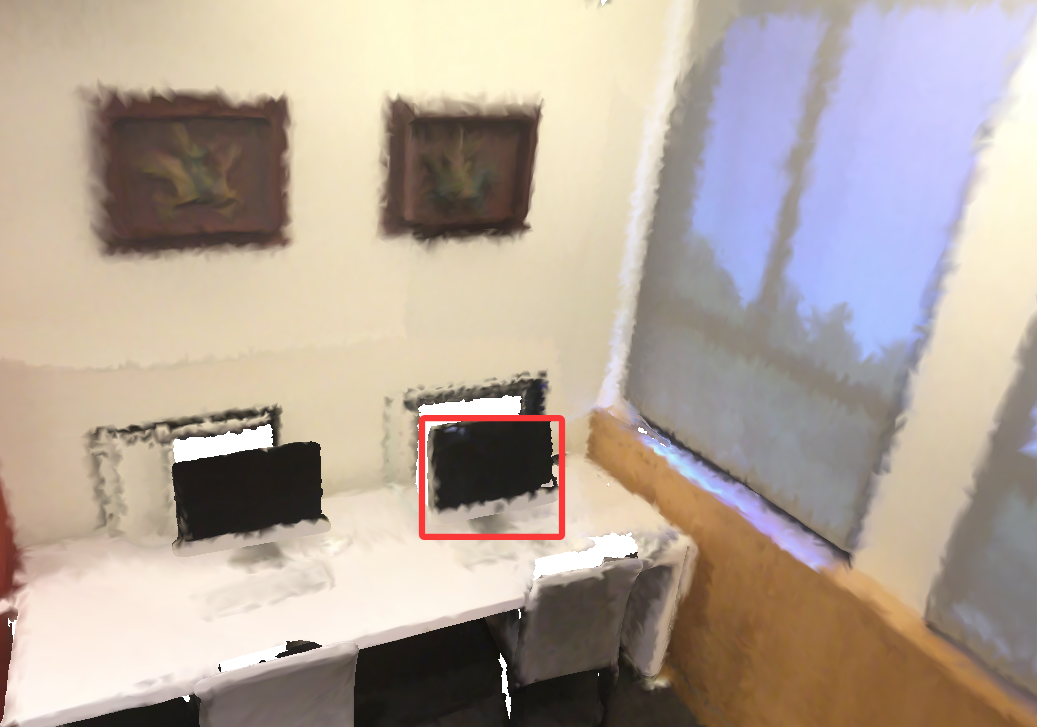}
    \end{minipage}\hfill
    
    \vskip 0.15in
    \hrule
    \vskip 0.15in
    
    \begin{minipage}{0.28\textwidth}
        The bookshelf is between another bookshelf and a red wall. The bookshelf is brown and rectangular.
    \end{minipage}
    \begin{minipage}{0.35\textwidth}
        \centering
        \includegraphics[width=0.98\linewidth]{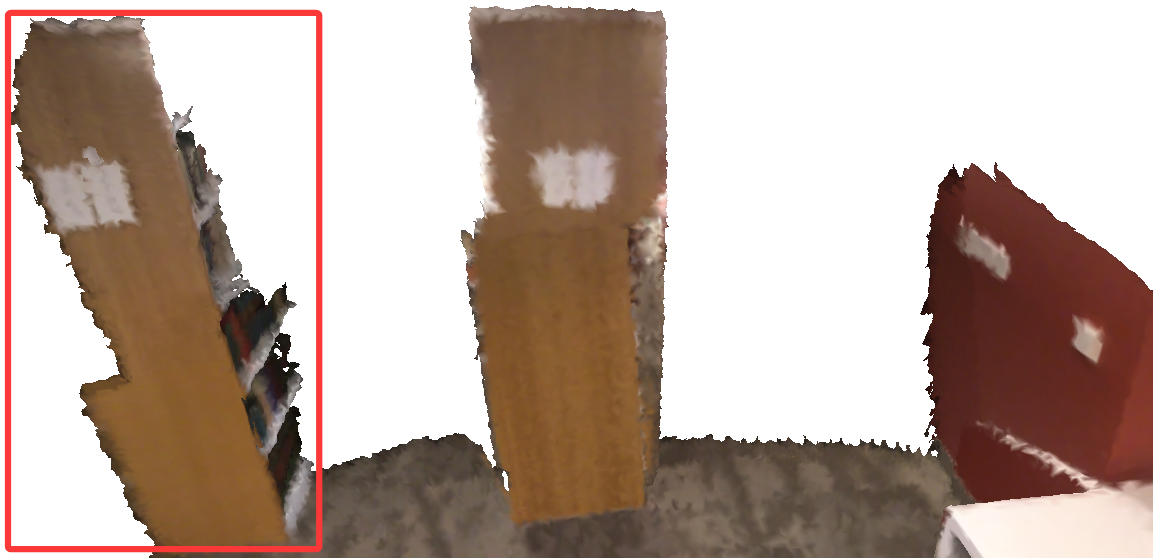}
    \end{minipage}
    \begin{minipage}{0.35\textwidth}
        \centering
        \includegraphics[width=0.98\linewidth]{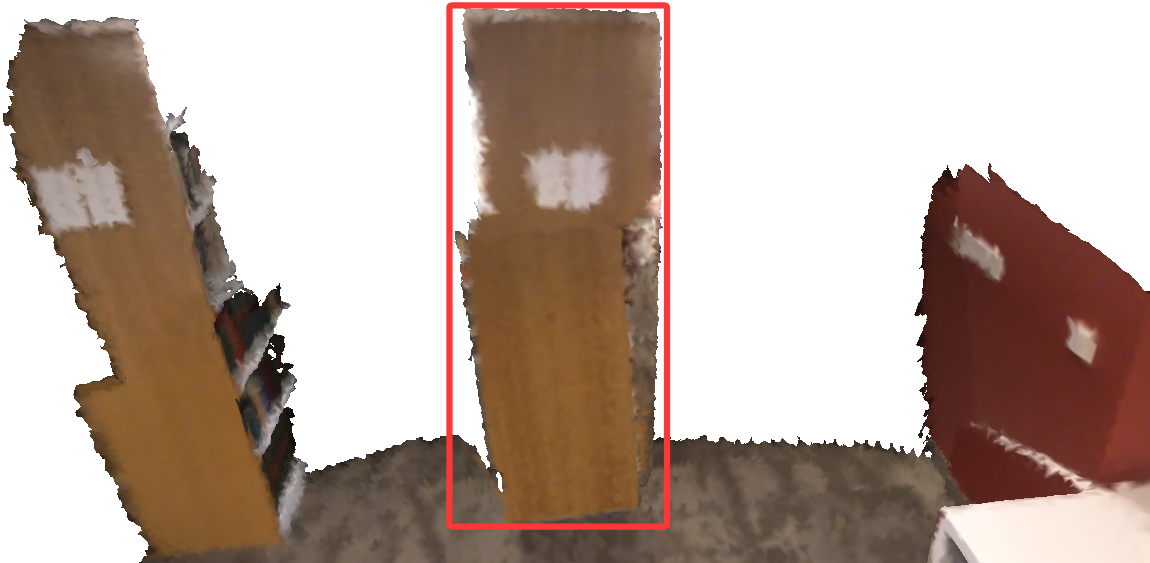}
    \end{minipage}\hfill
    
    \vskip 0.15in
    \hrule
    \vskip 0.15in
    
    \begin{minipage}{0.28\textwidth}
        The Ottoman is in the back, middle of the room. There is an identical ottoman to the right of it.
    \end{minipage}
    \begin{minipage}{0.35\textwidth}
        \centering
        \includegraphics[width=0.98\linewidth]{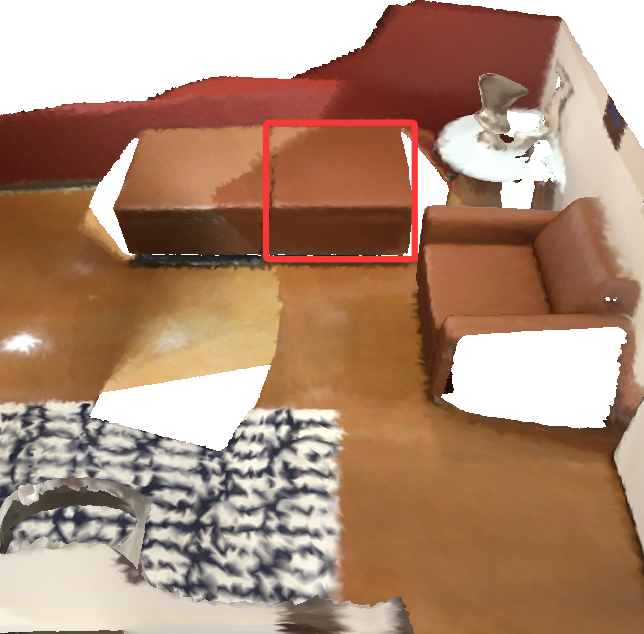}
    \end{minipage}
    \begin{minipage}{0.35\textwidth}
        \centering
        \includegraphics[width=0.98\linewidth]{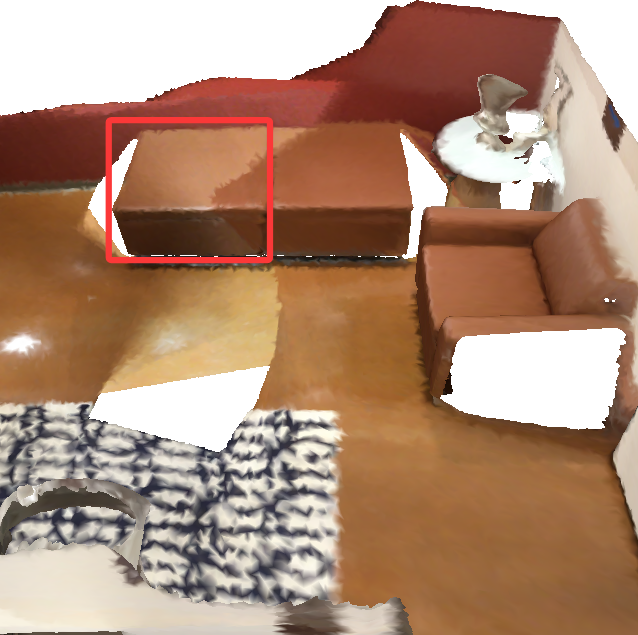}
    \end{minipage}\hfill

    \vskip 0.15in
    \hrule
\end{table*}

In this section, we provide additional qualitative results to illustrate the effectiveness of QuatRoPE. The qualitative results are obtained from Chat-Scene-1B's \cite{chat-scene} predictions on the validation split of the ScanRefer dataset \cite{scanrefer}.

The cases in Tables \ref{tab:qr1} - \ref{tab:qr3} demonstrate that QuatRoPE can effectively provide precise relative positions between objects.
By providing explicit spatial relations between objects, models can directly perceive the scene layout without extracting and calculating objects' positions from prematurely fused features.
Such a method significantly reduces the cost of training models to learn spatial reasoning, enabling them to achieve better performance.

\end{document}